\documentclass[manuscript,screen,nonacm]{acmart}
\setcopyright{none}
\renewcommand\footnotetextcopyrightpermission[1]{}
\acmConference{}{}{}
\acmBooktitle{}
\acmDOI{}
\acmISBN{}
\AtBeginDocument{%
  }

\setcopyright{acmlicensed}
\copyrightyear{2018}
\acmYear{2018}
\acmDOI{XXXXXXX.XXXXXXX}
\acmConference[]{Make sure to enter the correct
  conference title from your rights confirmation email}{June 03--05, 2018}{Woodstock, NY}
\acmISBN{978-1-4503-XXXX-X/2018/06}

\usepackage{enumitem}
\usepackage{tabularx}
\usepackage{booktabs}
\usepackage{multirow}
\usepackage{array}
\usepackage{caption}
\usepackage{subcaption}
\usepackage{multicol}
\usepackage{placeins}

\begin{document}

\title[Content Moderation Does Not Capture Communities' Heterogeneous Attitudes Towards Reclaimed Language]{IYKYK (But AI Doesn't): Automated Content Moderation Does Not Capture Communities' Heterogeneous Attitudes Towards Reclaimed Language}

\author{Christina Chance}
\thanks{Code available at \url{https://github.com/christinachance/reclaimed-language-community-attitudes}}
\email{christinachance315@gmail.com}
\orcid{0000-0002-8254-0670}
\affiliation{%
  \institution{University of California Los Angeles}
  \city{Los Angeles}
  \state{California}
  \country{USA}
}

\author{Rebecca Pattichis}
\affiliation{%
  \institution{Independent Researcher}
  \city{Albuquerque}
  \state{New Mexico}
  \country{USA}}
\email{}

\author{Arjun Subramonian}
\affiliation{%
  \institution{University of California Los Angeles}
  \city{Los Angeles}
  \state{California}
  \country{USA}
}

\author{James He}
\affiliation{%
  \institution{University of California Los Angeles}
  \city{Los Angeles}
  \state{California}
  \country{USA}
}

\author{Shruti Narayanan}
\affiliation{%
  \institution{University of California Los Angeles}
  \city{Los Angeles}
  \state{California}
  \country{USA}
}

\author{Saadia Gabriel}
\affiliation{%
  \institution{University of California Los Angeles}
  \city{Los Angeles}
  \state{California}
  \country{USA}
}

\author{Kai-Wei Chang}
\email{kwchang@cs.ucla.edu}
\affiliation{%
  \institution{University of California Los Angeles}
  \city{Los Angeles}
  \state{CA}
  \country{USA}
}

\renewcommand{\shortauthors}{Chance et al.}

\begin{abstract}
Reclaimed slur usage is a common and meaningful practice online for many marginalized communities. It serves as a source of solidarity, identity, and shared experience. However, contemporary automated and AI-based moderation tools for online content largely fail to distinguish between reclaimed and hateful uses of slurs, resulting in the suppression of marginalized voices. In this work, we use quantitative and qualitative methods to examine the attitudes of social media users in LGBTQIA+, Black, and women communities around reclaimed slurs targeting our focus groups including the f-word, n-word, and b-word. With social media users from these communities, we collect and analyze an annotated online slur usage corpus. The corpus includes annotators' perceptions of whether an online text containing a slur should be flagged as hate speech, as well as contextual features of the slur usage (e.g., whether derogatory, target of the slur). Across all communities and annotation questions, we observe low inter-annotator agreement, indicating substantial disagreement among in-group annotators. This is compounded by the fact that, absent clear contextual signals of identity and intent, even in-group members may disagree on how to interpret reclaimed slur usage online. Semi-structured interviews with annotators suggest that differences in lived experience and personal history contribute to this variation as well. We find poor alignment between annotator judgments and automated hate speech assessments produced by Perspective API. We further observe that certain features of a text such as whether the slur usage was derogatory and if the slur was targeted at oneself are more associated with whether annotators report the text as hate speech. Together, these findings highlight the inherent subjectivity and contextual nature of how marginalized communities interpret slurs online. They challenge assumptions that community membership alone is sufficient to establish reliable gold standards for content moderation. Results emphasize that context-sensitive approaches are critical to address online slur usage without over-moderating and further marginalizing online communities. 
\end{abstract}



\maketitle
\section{Introduction}

\begin{quote}
   \textit{Ultimately, struggles involving the N word and other forms of toxic language become intensely personal conflicts, waged and decided within our individual selves. Alone with our thoughts, impulses, and emotions, we are at liberty to weigh the arguments and make a choice at a protective remove from the clamor and heat of Orwellian crusaders. The primacy of individual choice and the esteem with which we Americans regard freedom of expression complicates our attitudes toward the N word.}
   \\\hfill--- Jabari Asim (2007)  \citep[Chapter~15, p.~229]{asim2007n}
\end{quote}

Social media platforms are central sites for community formation, cultural expression, and political engagement \cite{doi:10.1177/13548565241296628, doi:10.1177/1527476413480247, guta2015veiling, Theocharis07062023,Schuster01062013}. By enabling interaction across space and time, these platforms allow users to organize around shared identities, experiences, and values. However, alongside these affordances, platforms rely heavily on automated and community-based content moderation to govern user behavior. For marginalized communities, these systems often introduce a different constraint: over-regulation and suppression. 

Prior work documents how moderation systems disproportionately police identity-specific topics, enforce colonial and Western norms, and fail to capture the social and cultural nuance of marginalized communities \cite{Harris_2023, Shaid2023}. One particularly salient example of this failure is the treatment of reclaimed language. Reclaimed slurs are terms historically used to oppress marginalized groups that have been reappropriated as markers of identity, solidarity, and shared experience \cite{https://doi.org/10.1111/papq.12403, Reclamation, Pavone+2024+349+363}. While reclaimed language plays an important social role within communities, content moderation systems and hate speech detection models frequently flag these terms as abusive, reflecting a broader inability to account for context, intent, and community-specific meaning.

In response, other work has attempted to improve moderation performance by increasingly incorporating annotators who share the identity of the targeted community, under the assumption that in-group membership provides more faithful judgments.
However, even with more reliable labels, during training and testing models still have poor performance on differentiating slur usage in the context of toxicity detection \cite{dorn2024harmfulspeechdetectionlanguage}.
We hypothesize that this failure stems from a deeper assumption: that marginalized communities are internally homogeneous in their beliefs, experiences, and interpretations of reclaimed language. In practice, reclaimed slur usage is deeply personal and shaped by lived experience, including familial, cultural, and environmental factors. While in-group perspectives are necessary for understanding harm, they do not guarantee consensus. Disagreement is not an anomaly in these settings but a defining feature of how reclaimed language is interpreted.

This challenge is further compounded by limitations in annotation practices. Many hate speech datasets lack sufficient contextual information to capture speaker intent, audience, or relational dynamics, all of which are critical to interpreting reclaimed slur usage. As a result, reclaimed language occupies a difficult and underexplored space in hate speech detection research, sitting at the intersection of language, identity, community norms, and moderation governance.

In this study, we examine how members of marginalized communities interpret reclaimed slur usage and decide whether such content should be reported as hate speech. We focus on LGBTQIA+, Black, and women communities, examining the use of the f-word, n-word, and b-word as reclaimed language within these groups. From a corpus of approximately 12,000 tweets containing slurs drawn from Twitter and existing hate speech datasets, we select 100 texts per slur for annotation. Annotators label features of each text including whether the slur is used in a reclaimed manner, the target of the slur, the surrounding context, and whether the content should be reported as hate speech. We complement this annotation task with semi-structured interviews to examine how personal experiences with reclaimed language shape their interpretations of slurs and online behavior. 
Through this mixed-methods approach, we interrogate assumptions of intra-community homogeneity, examine how lived experience influences interpretations of reclaimed slurs, and analyze how norms vary across different reclaimed terms. 

We center our analysis around the following research questions:
\begin{enumerate}
    \item [RQ1.] \textit{To what extent does disagreement within identity groups challenge prior work that treats identity as a reliable predictor of labeling decisions?}
    \item [RQ2.] \textit{What are the most predictive features of whether someone thinks that content should be flagged as hate speech?}
    \item [RQ3.] \textit{Does model calibration reflect real world uncertainty for flagging content?}
    \item [RQ4.]  \textit{What patterns and differences emerge in how reclaimed slurs are interpreted across different marginalized communities?}
\end{enumerate}

Our results indicate that shared identity group does not suggest improved agreement-wise reliability, as we saw low Krippendorff's alpha agreement for all annotation questions for each community (f-word $\alpha = 0.15$, n-word $\alpha=0.06$, b-word $\alpha=0.21$). The agreement is only marginally higher for reporting the text as hate speech if the author of the tweet is in-group (in-group $\alpha=0.18$, out-group $\alpha=0.15$) [\textit{RQ1}]. Features like whether the text is derogatory as well as who the target is influences the overall decision for reporting, with features like the type of reclamation and the salient context of the slur usage influencing the difference in reporting for in-group vs. out-group authorship [\textit{RQ2}]. When looking at the alignment of whether the annotator would report the text to the output of Perspective API, we find that the model aligns more with the assumption that the author is out-group, having overall lower average total variation (ATV) for out-group authorship compared to ingroup authorship, with the exception of the b-word 
[\textit{RQ3}]. Additionally, we find that when comparing salient feature influence and alignment with model outputs, each slur shows unique interpretation patterns suggesting, distinct community attitudes to slur usage. For example, the usage of the slur in the context of ``discussion of slur'' increases the likelihood for change in reporting based on authorship for the n-word, while for the f-word that feature actually reduces the likelihood that the label changes [\textit{RQ4}]. These results suggest a need for more nuanced, community-informed, and flexible definitions of hate speech as it relates to online language and behavior of marginalized communities.

\section{Literature Review}
We position this work as a discussion of oppression through the lens of social media content moderation and, more broadly, automated NLP systems' inability to contextually understand marginalized communities. We highlight the concern and harm that results in homogenizing communities via shared beliefs and preferences. Annotator disagreement, especially in subjective and personal tasks like hate speech detection, is expected within a community. We provide grounding for this paper in relation to existing works and discussions (see extended literature review in Appendix~\ref{appendix:literature_extended}).

\noindent\textbf{Reclaimed Slurs.}
Slurs are pejorative terms targeting social groups (e.g., race, ethnicity, nationality, religion, sexual orientation, gender, disability). Historically rooted in oppression within the context of the U.S., they have been used to demean communities and elevate the speaker's social status. While socially taboo, some target communities have reclaimed these words to signal resilience, pride, and solidarity. Reclamation has been conceptualized in multiple ways, including distinctions between casual versus socio-political use \cite{BIANCHI201435}, levels of individual and group adoption \cite{Galinsky}, and dynamics of power and self-stigmatization \cite{Reclamation}. For this study we adopt \citet{jeshion}'s polysemy model, which differentiates \textit{insular} reclamation (in-group camaraderie) and \textit{pride} reclamation (expressing dignity and honor in group membership).

\noindent\textbf{Automated Content Moderation Suppression.}
Automated moderation disproportionately suppresses posts by marginalized users by over-enforcing policies on identity-related speech. Prior work shows that Black, trans, and gender-queer users experience disproportionate content removal, shadow banning, and inconsistent enforcement under vague or Western-centric norms \cite{Harris_2023, Shaid2023, Haimson2021}. Models and datasets also amplify these disparities, often flagging non-derogatory reclaimed language as harmful while overlooking context and lived experience \cite{davidson-etal-2019-racial, vidgen-etal-2021-learning, hartvigsen-etal-2022-toxigen, dorn2024harmfulspeechdetectionlanguage}.  

Our work builds on \citet{dorn2024harmfulspeechdetectionlanguage}, expanding annotation methodology to examine salient context, slur usage, and author group membership across multiple communities, highlighting how moderation decisions intersect with identity beyond the gender-queer population.

\noindent\textbf{Reclaimed Language in NLP.}
Although hate speech detection has been widely studied, reclaimed language remains understudied. Most work addresses it indirectly through identity term bias \cite{attanasio-etal-2022-entropy, 10.1145/3278721.3278729, sap-etal-2019-risk} and keyword bias studies \cite{YIN2022100210, cercas-curry-etal-2024-subjective}. Few studies directly examine reclaimed language. Prior work proposes taxonomies of derogatory, reclaimed, and counter-speech slur usage and documents annotation challenges and Perspective API bias \cite{kurrek-etal-2020-towards}. Other studies show that hate speech models disproportionately flag in-group LGBTQIA+ reclaimed slur usage as harmful, increasing false positives \cite{dorn2024harmfulspeechdetectionlanguage,10.1145/3614419.3644025}. While some approaches fine-tune models to predict reappropriation using homo-transphobic datasets \cite{draetta-etal-2024-reclaim}, they continue to frame reclamation as a binary classification problem. Overall, existing NLP work treats reclaimed language primarily as a modeling challenge rather than a socially situated practice shaped by context and lived experience.

\noindent\textbf{Annotator Disagreement.}
Annotator identity, including demographics, lived experiences, and beliefs, significantly shapes subjective labeling tasks such as toxicity and hate speech detection \cite{sap-etal-2022-annotators, goyal2022toxicitytoxicityexploringimpact, pei-jurgens-2023-annotator, biester-etal-2022-analyzing}. This variance manifests across multiple dimensions of rater identity: \citet{al-kuwatly-etal-2020-identifying} train models on the same data with distinct annotator groups and find significant performance differences along lines of first language, age, and education, while \citet{waseem-2016-racist} find that amateur annotators label more text as hate speech compared to experts, even when both groups receive training. Cross-cultural studies further highlight this instability such as \citet{lee-etal-2024-exploring-cross} who found substantial label disagreement across countries, driven by differing interpretations of sarcasm and personal biases on divisive topics. Compounding this, \citet{santy-etal-2023-nlpositionality} show that annotators with different backgrounds align with different models, and that datasets and models disproportionately reflect the perspectives of white, Western, college-educated annotators. Simple aggregation of labels or in-group versus out-group approaches \cite{fleisig-etal-2023-majority, 10.1145/3555088} fail to capture the nuanced, heterogeneous perspectives within identity groups and are therefore insufficient to fully represent lived experiences and interpretive differences.

Recent work models annotator differences via embeddings, demographic or historical information \cite{deng-etal-2023-annotate, 10.1609/aaai.v37i12.26698}, distribution-aware calibration \cite{baan-etal-2022-stop}, and modular pluralism frameworks for LLM alignment \cite{feng-etal-2024-modular, bansal2025comparingbadapplesgood}, producing models sensitive to heterogeneous judgments rather than enforcing a single ``gold standard''. Such research underscores that annotator disagreement reflects the socially situated nature of language interpretation, not noise.

\section{Methods}
Our study aims to highlight the impact of lived experiences, identity, and community influence on the use of reclaimed language associated with three marginalized communities: Black, LGBTQIA+, and women communities. We show that one-size fits all approach to handling reclaimed language within each community is problematic, can further suppress marginalized voices, and assumes homogeneity in use and acceptability. We do this through the development of a corpus with examples of online reclaimed language use and in-group (i.e. community member) annotations of perceptions of slur use.
Additionally, we conduct semi-structured interviews seven annotators, learning about how their lived experiences and environment have shaped their understanding and use of reclaimed words.

\subsection{Studied Communities and their Slurs}
We focus on three marginalized communities: Black, LGBTQIA+, and women. These communities have reclaimed slurs (e.g., the n-word, f-word, b-word) that have been historically used to oppress them. Note that these slurs may only be reclaimed by a subset of each community, there are other slurs used historically for each community, and some slurs (e.g., the b-word) have been reclaimed across communities. We selected these general groups as to not identify or assign a specific word to a specific group. As these are all communities we care about, we wanted to ensure alignment through bounding the annotators as to be members of these specific groups. We examine these groups because the social trajectories of their respective reclaimed words reveal how language, power, and visibility interact in public discourse.

For example, the b-word has become widely normalized in contemporary culture \cite{https://doi.org/10.1111/papq.12403, zhou2020voxsuffrage}. It appears not only in casual conversation but also in marketing campaigns (e.g., Wiggles's ``SheIsAB*tch" campaign \cite{mediabrief2021sheisabitch}), restaurant branding (e.g., ``Bacon B*tch'' and ``Breakfast B*tch'' \cite{hatchett2024bitch}), and product names (e.g., ``B*tchstix'' lip balm \citep{bitchstix}). People of all gender identities use this word across contexts to denote both camaraderie and hate, reflecting its full integration into mainstream American culture. In contrast, the n-word, reclaimed by the Black community, remains a taboo term both socially and professionally. Its use continues to provoke debate about who can and cannot say it, with public incidents often leading to social backlash, firings, and other consequences \cite{apnews2025racistslur, nypost2026njcops}. Despite its reclamation, the word retains immense power, frequently surfacing in discussions of race, censorship, and accountability, while still being weaponized by white supremacists and nationalists \cite{alt-right, Bradley2023}.
Similarly, the f-word carries enduring stigma. During the early 2000s, mainstream culture normalized casual homophobia, making the word appear socially acceptable in entertainment and everyday speech. However, advocacy and heightened queer visibility in the mid-2010s prompted public reevaluation, reinforcing the recognition of this word's harm and ushering in an era of accountability \cite{brammer2019lgbtqvisibility, allegretti2023vicehomophobia}. Taken together, these histories illustrate that the social acceptability of reclaimed language is not static but instead reflects broader structures of power, cultural visibility, and resistance. Words reclaimed by marginalized groups are accepted or rejected by the mainstream in ways that mirror whose voices are recognized, amplified, or silenced within society.

\subsection{Participant Recruitment and Demographics}
To recruit members of the Black, LGBTQIA+, and women communities, each main author shared the research project through Black in AI\citep{blackinai}, Queer in AI \citep{QAI2023}, and Women in ML\citep{wiml} channels. We recruited within these channels to control for  demographic features like education and understanding of the AI, NLP, and content moderation research space. Because this is such a specific subgroup of people, we do not claim generalizability for the findings. Also to note, as it is such a diverse sample of individuals, some participants are English as Second Language (ESL) speakers, but we do not control for this and discuss the implications within the discussion. Additionally, to reduce annotator bias, we opted not to provide training, as we believe the population we are sampling from already understands the importance of quality annotations.

For each studied slur, we use a pre-screening survey, as seen in appendix~\ref{appendix:prescreen_survey}, to get our sample population ($n_{n-word} = 6; n_{f-word}= 6;$ $n_{b-word}=9$). Prior empirical work shows that small samples are sufficient to capture the relevant variation in focused, well-defined annotation tasks \cite{HENNINK2022114523}, which motivates the sample size used in this study. Via our pre-screening survey, we collect fine-grained demographic information within the categories of race and ethnicity, pronouns, and sexuality. We model the categories after the 2022-2023 Queer Experiences in AI Survey hosted by Queer in AI\citep{qai-survey}, and extend some categories. We allow survey respondents to select which reclaimed word they would like to annotate, since they may exist at the intersection of multiple marginalized communities. Through this pre-screening, if a participant does not belong to a community that has reclaimed a slur or is not comfortable with the task they are filtered out. We additionally ask about their perceptions and usage around  in- and out-group  members for the reclaimed word they choose to annotate. At any time in the annotation process, participants are able to end their participation and withdraw from the study if they would not like to continue.

This study was deemed by the Institutional Review Board (IRB) as IRB exempt. Through the prescreening we received 54 submission for participation in the annotation process. Of the 54, 50 participants said they were willing to be interviewed. The prescreening was open from October 2024 to April 2025. The goal was to have 15 participants for each of the reclaimed words, therefore we sent out 49 emails, as 5 people who completed the survey noted they did not to participant in the annotation process, inviting participants to perform the task. During this step an additional 8 participants opted out of continuing in which we had a final count of 41 email sent that were either not responded to or said they would do it ($n_{n-word} = 11; n_{f-word}= 16;$ $n_{b-word}=14$). We received 21 completed annotation tasks ($n_{n-word} = 6/11; n_{f-word}= 9/16;$ $n_{b-word}=6/14$) in which they were compensated with a \$45 virtual Amazon gift card. For the interviews, we emailed all participants that completed the task and stated in their email that they were still interested in being interviewed. We sent emails officially inviting 16 annotators to participate in interviews and completed 7 total interviews ($n_{n-word} = 2; n_{f-word}= 3;$ $n_{b-word}=2$) in which they were compensated with a \$15 virtual Amazon gift card. Overall the drop out rate from pre-screening to completion of task was $61.1\%$.

A main focus of this work is to debunk the idea that a subsample of a diverse and robust population is enough to represent the belief system, opinions, and preferences of the community. There is no singular belief system or shared opinion that a community holds, especially on topics so subjective and personal such as hate speech and slur usage, in which context, social standing, and lived experiences are foundational to a person's understanding. Hence, ``representativeness'' is not a goal of this study nor for the recruitment of the subpopulation annotating and participating in interviews.
We sought people that were willing and able to provide their insight, experiences, and time.

\subsection{Dataset Design}

\noindent\textbf{Raw Data Collection.}
Using the X Developer API, we gather approximately 12,000 posts that contain at least one of the b-word, f-word, and n-word and common variations. We additionally look into existing hate speech and toxicity datasets including Dynahate \cite{vidgen-etal-2021-learning}, Toxigen \cite{hartvigsen-etal-2022-toxigen}, and Jigsaw Specialized Rater Pools \cite{10.1145/3555088} and extract a subset of examples that contain reclaimed language.
We pre-process all examples to remove URLs, de-identify X handles, and convert emoticons to their textual descriptions. We further combine all data for each reclaimed word and, similarly to \citet{dorn2024harmfulspeechdetectionlanguage}, select the texts that were judged as highly toxic and not toxic based on the identity attack axis using Detoxify \cite{Detoxify}.

\noindent\textbf{Annotation Process.}
We design the annotation task to assess \textit{how} annotators perceive reclaimed slurs to be used in online text through a set of predefined descriptors. We additionally ask whether the annotator personally considers the content of the text to be hate speech. The questions and options are listed below. We provide the actual text given to the annotators in Appendix Table~\ref{appendix:annotation_instruct}.

\begin{enumerate}

       \item [\textbf{Q1.}]\textbf{Suppose the author is in-group. What kind of reclamation is present in the use of the slur in this context?} Here, we use the proposed framework of \citeauthor{jeshion} that defines two use cases of reclaimed language,
    \begin{itemize}
        \item \textbf{Pride reclamation}: the use of slur as an expression of pride for being in-group, which is presented as an acceptable manner of referencing the group (e.g., queer and black).
        \item \textbf{Insular reclamation}: the use of a slur as an expression of camaraderie among the members of a group, in which it is presented as an unacceptable manner of referencing the group (e.g., n-word and b-word).
    \end{itemize}
     and we include ``neither'' as a possible option to indicate that the slur is not being used in reclaimed manner. The words are defined below as they are defined in the annotation instructions provided. (\textit{Considered in RQ1})
  
    \item [\textbf{Q2.}] \textbf{Do not make assumptions about whether the author is in-group or out-group. Is this a derogatory use of the word?} This question is a binary {\em yes} or {\em no} question. (\textit{Considered in RQ1 \& RQ2})

    \item [\textbf{Q3.}] \textbf{Whom is the slur directed at?}  We provide the following options:  oneself, another individual, individuals from a community associated with the slur, individuals not from a community associated with the slur, a known subset of people, a broader group of people. (\textit{Considered in RQ1 \& RQ2})

    \item [\textbf{Q4/5.}] \textbf{What is a salient/secondary context in which the word is being used?} We apply the slur usage taxonomy from \citet{kurrek-etal-2020-towards} and subcategories and definitions from \citet{dorn2024harmfulspeechdetectionlanguage} to categorize the usage type of each slur: counter speech, quote, discussion of slur, discussion of identity, sexualization, sarcasm, recollection, neologism. We provide definitions for these categories in Appendix \ref{appendix:annotation_instruct}. The secondary context question (Q5)  was not required. (\textit{Considered in RQ1 \& RQ2})

    \item [\textbf{Q6/7.}] \textbf{Suppose the author is in-group/out-group. Would you want a content moderation model to report this as hate speech?} Annotator considers both counterfactual -- if the annotator is in-group (Q6) and if the annotator is out-group (Q7). This question is a binary {\em yes} or {\em no} question. (\textit{Considered in RQ1, RQ2, \& RQ3}) 
\end{enumerate}

We provide further discussion on the framing and purpose of each annotation question in Appendix ~\ref{appendix:annotation_questions}. We gave participants around two weeks to complete the annotation task, encouraging them to take breaks and consider their mental health, as hateful and offensive language is present in the the data.
If annotators were unsure of the answer to a question, they were encouraged to select an answer and if not possible, to leave the question blank.
 
\subsection{Interview Protocol}
We conduct semi-structured interviews to learn about how annotators' lived experiences, cultural upbringing, and communities influence their understanding of reclaimed language use. See Appendix \ref{appendix:interview_questions} for interview questions. While the annotations allow us to quantify differences in perception, many qualitative experiences around reclaimed language cannot be quantitatively captured.
As we discuss in Section~\ref{sect:discussion}, our interviews suggest that people in the same community have varying perspectives on reclaimed language use. Both the annotation of the corpus as well as the interviews rely on a limited and non-representative sample. While annotators may not reflect the full diversity of perspectives within the communities studied, the divergences in their perspectives are meaningful.

\section{Discussion}\label{sect:discussion}
\subsection{RQ1: To what extent does disagreement within identity groups challenge prior work that treats identity as a reliable predictor of labeling decisions?}

\begin{table}[ht]
\small
\centering
\scalebox{0.9}{
\begin{tabular}{lrrrrrrr}
\toprule

 & \multicolumn{1}{c}{\begin{tabular}[c]{@{}c@{}} Reclaimed\\ (Q1) \end{tabular}} & \multicolumn{1}{c}{\begin{tabular}[c]{@{}c@{}}Derogatory\\ (Q2) \end{tabular}} & \multicolumn{1}{c}{\begin{tabular}[c]{@{}c@{}}Target\\ (Q3) \end{tabular}} & \multicolumn{1}{c}{\begin{tabular}[c]{@{}c@{}}Context\\ (Q4)\end{tabular}} & \multicolumn{1}{c}{\begin{tabular}[c]{@{}c@{}}Sec. Context\\ (Q5)\end{tabular}} & \multicolumn{1}{c}{\begin{tabular}[c]{@{}c@{}}Report\\In-group (Q6) \end{tabular}} & \multicolumn{1}{c}{\begin{tabular}[c]{@{}c@{}}Report\\ Out-group (Q7) \end{tabular}}\\
 \midrule
F-Word &  0.07 & 0.27 & 0.16 & 0.13 & \textbf{0.02} & 0.17 & 0.24\\
B-Word &  \textbf{0.18} & \textbf{0.33} & \textbf{0.24} & \textbf{0.22} & 0.01 & \textbf{0.20} & \textbf{0.25}\\
N-Word & 0.05 & 0.19 & 0.19 & -0.03 & -0.07 & 0.17 & -0.06\\

\bottomrule
\end{tabular}
}
\caption{\label{tab:krippendorf_alpha} Krippendorff's alpha annotator agreement score for each annotation question. Values range from -1 to 1, where 1 indicates perfect agreement, 0 indicates random chance agreement, and negative indicates systematic disagreement. Bolded values are the highest agreement per question.}
\end{table}

In Table \ref{tab:krippendorf_alpha}, we compute Krippendorff's alpha agreement score \cite{castro-2017-fast-krippendorff} for each annotation by slur. No agreement scores surpass $\alpha = 0.33$ ($\alpha = 0.80$ is considered high \cite{MARZI2024102545}). The highest agreement across all communities is for Q2 (whether text is derogatory), suggesting that there may be broader agreement on whether text is intended to be disparaging. Besides Q5, which is optional, Q1, which asks about the type of reclamation, has the lowest agreement across all communities, with $\alpha = 0.05$ for the n-word. Even when treating both types of reclamation as the same label, agreement scores were still extremely low (f-word $\alpha=0.096$, b-word $\alpha=0.211$,  n-word $\alpha=-0.025$). When treating reclamation types as distinct labels, full annotator agreement (i.e., all annotators agreeing, distinct from Krippendorff’s $\alpha$) was extremely low (3–7\%) and occurred almost entirely in the “neither” category, with no agreement on insular or pride cases. Collapsing these into a single “reclaimed” label slightly increased full agreement (11–13\%), driven by modest gains in reclaimed cases (e.g., up to 28.6\%), though overall agreement remained low. Annotators found this question difficult, as there were many instances where they could not imagine the word being used in a reclaimed manner in the given text. Some further noted that they could not easily distinguish between the types of reclamation in the context of the text. 

Q4/5 (salient context) were challenging for the annotators, as they are rooted in trying to understand the intended usage of the slur; however, context and intent are often lost through text \cite{doi:10.1177/1461444810365313, 10.5555/2845794.2845803}.
In a post-annotation conversation with $P2$, we found that when reasoning through an author's labeling for the tweet \textit{``RT @[MASK]: B*tches want a lemonade moment so bad [face\_with\_tears\_of\_joy] [face\_with\_tears\_of\_joy] [face\_with\_tears\_of\_joy] [face\_with\_tears\_of\_joy]''} for salient context, both people contextualized the meaning of the tweet differently, leading to opposite labels. While one person assumed this tweet had to do with B\'eyonce, the other one viewed it as nonsensical. 
This exemplifies how cultural background can cause divergent understandings of the intent of the same text between in-group members. Importantly, these disagreements are not uniform across reclaimed terms. Later cross-slur analysis (RQ4) shows that the sources of disagreement vary by slur, reflecting differences in how each term is socially embedded, contextualized, and normatively constrained.

Missingness of annotations also influenced overall agreement in which Krippendorff's alpha ignores empty cells. Some annotators, especially those who annotated for the b-word, ran into issues with labeling (e.g., the b-word was used to reference an actual dog). Similarly, for the f-word, some annotators noted several instances where it was used to refer to cigarettes in British and Australian slang $[P9, P10]$. 
One annotator noted that the b-word specifically was challenging as it had been adapted to be used as a verb, to describe someone complaining or whining about something $[P2]$. Another annotator who evaluated the b-word noted that it occurred in texts with other slurs or identifiers of specific identity groups; they did not feel comfortable annotating these texts as they were not of those groups $[P8]$. Further analysis showed that of the 300 annotated posts, 25 shared more than one slur among the three slurs we considered. Further study of the interplay between reclaimed words is needed. However, we did find that within this subset of 25 texts, 66.9\% of annotators labeled these tweets as being authored by an out-group member, compared to the entire corpus rate of 55.8\%. Similarly, 75.1\% labeled these texts as derogatory compared the the corpus rate of 58.5\%.
Overall, we found the following missingness rates (not including Q5, which was optional): n-word $23/3600$ $(0.64\%)$, b-word $631/5400$ $(11.69\%)$, and f-word $135/3600$ $(3.75\%)$.\footnote{Missingness with secondary salient context: n-word $402/4200$ $(9.57\%)$, b-word $1363/6300$ $(21.63\%)$, and f-word $611/4200$ $(14.55\%)$} These co-occurrence patterns suggest that reclaimed slurs often function within broader antagonistic or identity-laden discourse rather than as isolated lexical items, a trend that becomes especially pronounced for the b-word in our comparative analysis across slurs (RQ4).

\begin{figure}
    \centering
    \includegraphics[width=1.0\linewidth]{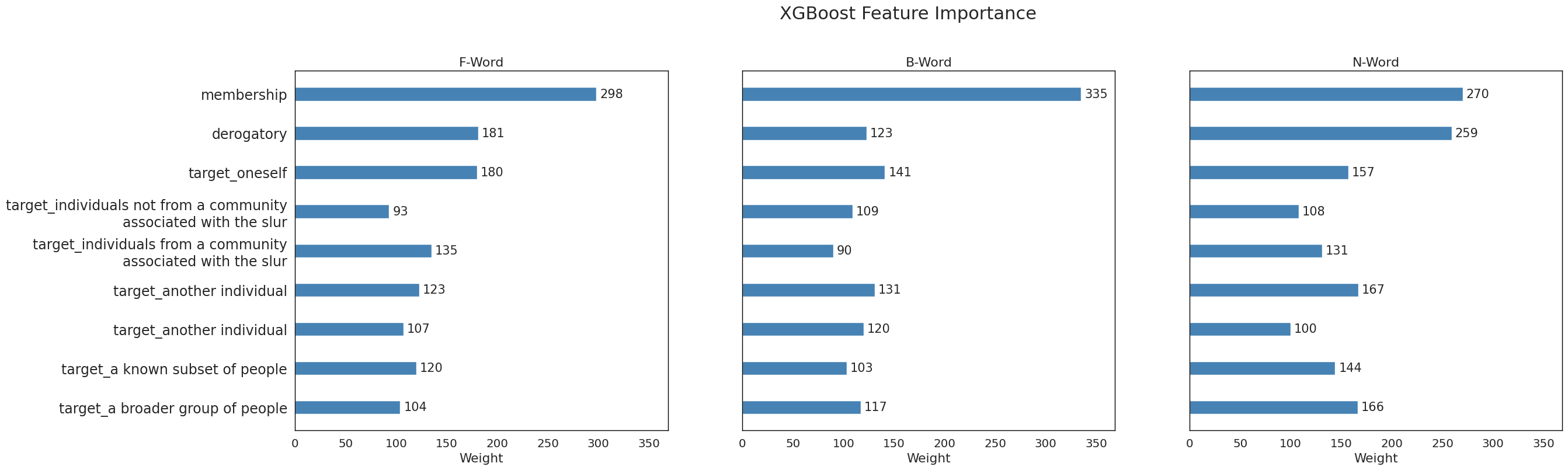}
    \caption{XGBoost feature importance measured by weight, showing the raw frequency each feature was used to split nodes across all trees. Features with more than two options were one-hot encoded into binary variables. We omit the N/A values for the primary salient context.}
    \label{fig:xgboost}
\end{figure}

We argue that disagreements in annotation are an essential characteristic of how reclaimed language is perceived within groups. Taken together, our findings show that disagreements are not incidental noise or annotator error, but a substantive signal of how meaning is negotiated within marginalized communities.
While annotators exhibited relatively higher agreement on more surface-level or seemingly objective features (e.g., identifying the target of a slur), agreement sharply declined for questions that involved interpretive judgment, emotional response, or personal boundaries around harm.
This pattern reflects the deeply personal nature of reclaimed language, where interpretation is shaped by lived experience, cultural exposure, comfort, and individual histories with the term itself.

\subsection{RQ2: What are the most predictive features of whether someone thinks that content should be flagged as hate speech?}

One point was consistent throughout our interviews: context matters. Various features need to be considered to better contextualize reclaimed slur use. Through these annotations, we assess which features of the text were most associated with participant's decision to flag text as hate speech. Specifically, we explore which features: (1) may most influence whether or not text is reported, and (2) influence the reporting given the group membership of the text author.

\begin{figure}
    \centering
    \includegraphics[width=1.0\linewidth]{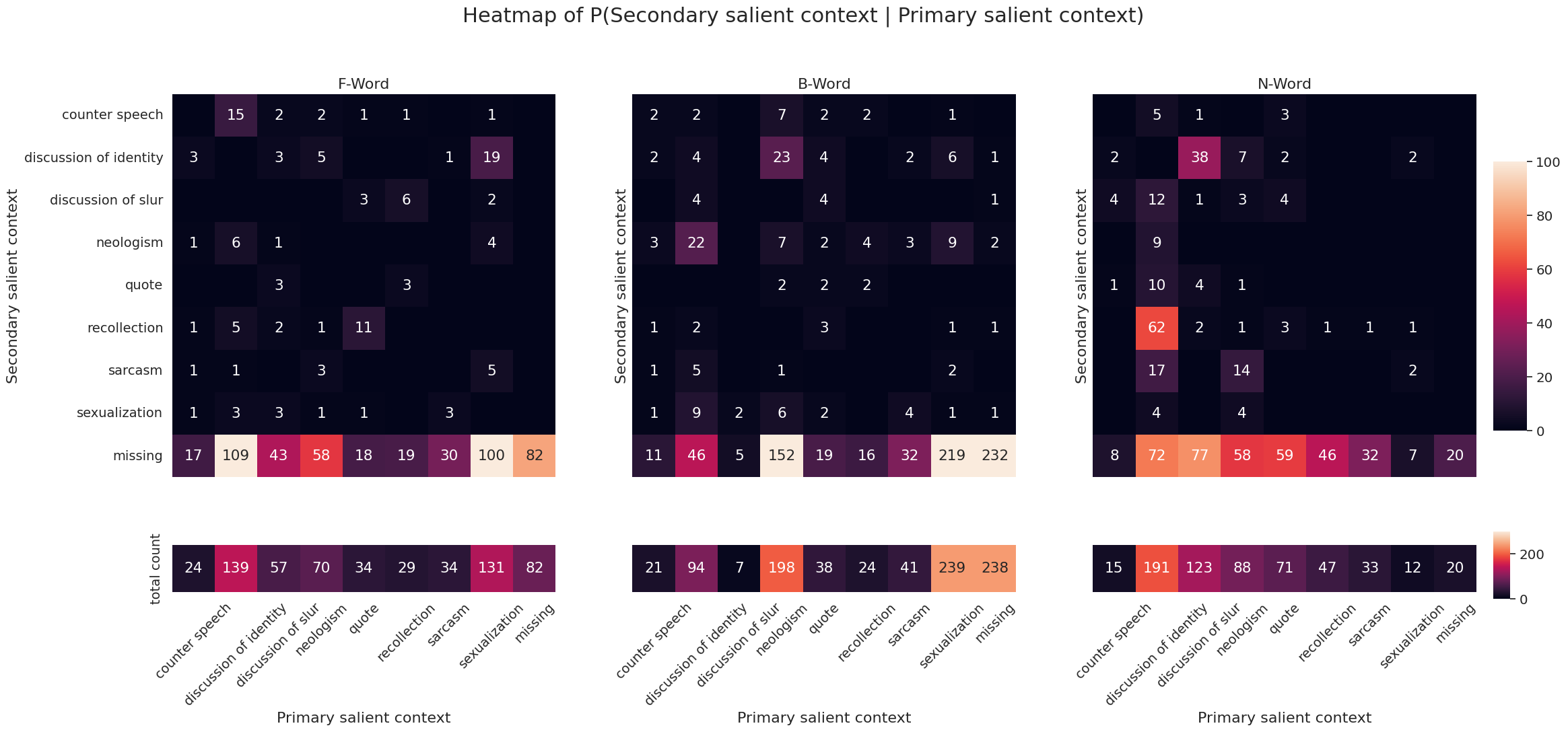}
    \caption{Heatmap depicting the raw annotation counts for a text's secondary salient context, disaggregated by its annotated primary salient context. The bottom row reports the total number of annotations for each primary salient context label.}
    \label{fig:context_heatmap}
\end{figure}

In Figure~\ref{fig:xgboost}, we present feature importance scores from an XGBoost model using the weight metric, which reflects how frequently a feature is used across all tree splits with larger values suggesting more influential features. To examine the influence of author group membership, we restructure the data by duplicating each annotated tweet: one version assumes an in-group and the other an out-group author. The two original reporting columns are merged into a single prediction label corresponding to the assumed authorship question response (Q6 \& Q7). Multi-option features are one-hot encoded, and we exclude the salient-context features, as we more closely examined the relationship of salient context (Q4) as a feature in Figure~\ref{fig:context_heatmap}. 

Group membership appears as the most frequently used feature in the model's decision process. As this feature is the only feature to vary across the duplicated records, this makes sense. The whether perceived as derogatory label (Q2) and cases where the slur targets oneself (Q3) also emerge as prominent features, suggesting that both perceived disparaging and self-directed targeting may play substantial roles in predicting whether annotators choose to report text as hate speech. However, the extent to which group membership functions as a dominant cue is not consistent across slurs. As we show in RQ4, the b-word and f-word exhibit stronger reliance on assumed author identity, while judgments around the n-word more heavily prioritize perceived derogation, indicating slur-specific normative decision pathways.

We also perform linear mixed-effects modeling on the annotation data, where the response variable is whether the group membership of the author impacts the decision to report the text (see Table~\ref{tab:mixed-effect} in the appendix).
Specifically, we map this value to 1 if the reporting decision changes between assumed in-group vs. out-group authorship and 0 if it does not.
Across all slurs, we see that annotators are more likely to change their reporting decision if the slur is used in a pride reclamatory manner. When the n-word appears in quoted contexts, annotators are much more likely to rely on author group membership to decide whether the content should be reported.
With regards to salient context, neologism, and quotes have a negative association with author group membership on reporting. 
When annotators label text as derogatory, there is not substantial alignment between author membership and reporting, suggesting that annotators tend to treat derogatory text similarly regardless of author's group membership.
These results also indicate that there are not distinct trends across the slurs in relation to whom the text targets. These trends suggest that annotators do not apply a single, consistent rule for evaluating reclaimed language; instead, different slurs may activate different interpretive ``heuristics,'' which RQ4 shows are tied to each term’s socialization, mainstreaming, and historical burden.
Unlike for the n-word, for the f-word, the target of the slur tends to somewhat influence changes in reporting. This aligns with our knowledge of how these slurs have been socialized. 
Due to the size of the data for each slur, our results are not statistically significant.

\subsection{RQ3: Does model calibration reflect real world uncertainty for flagging content?}

\begin{figure}
    \centering
    \includegraphics[width=\linewidth]{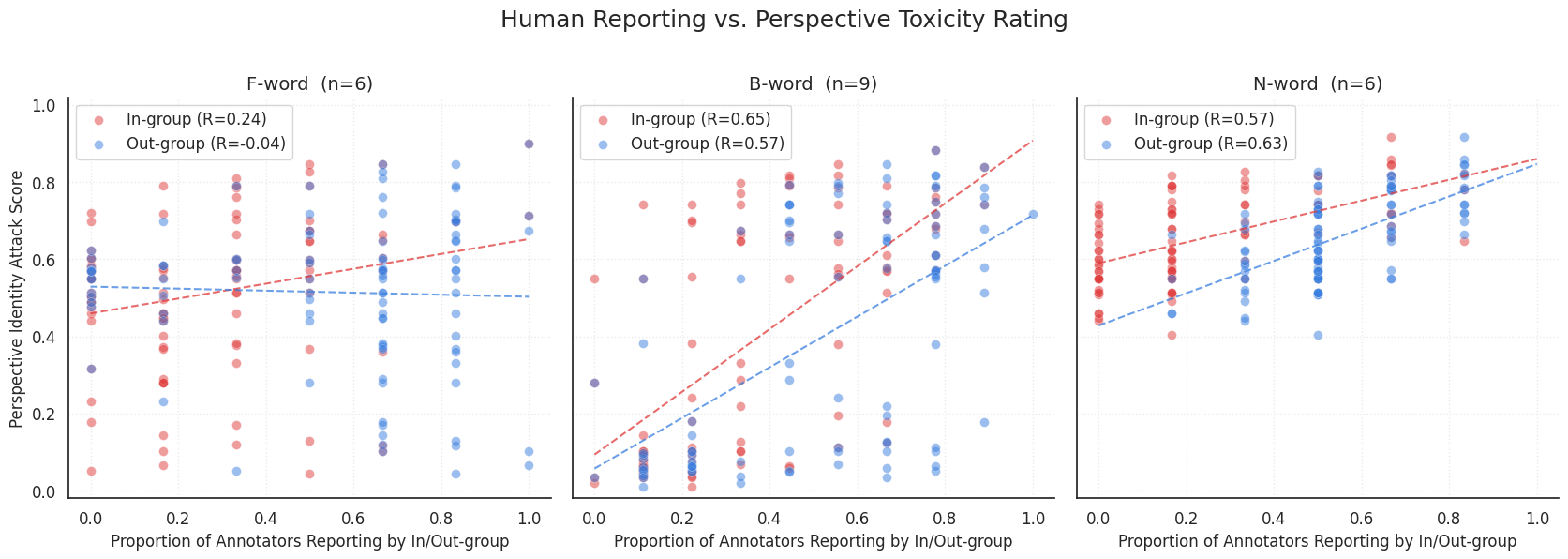}
    \caption{Alignment between human consensus and Perspective API varies by slur. Each tweet is represented as a pair of points sharing the same y-value (Perspective API identity attack score), since Perspective API does not differentiate between in-group and out-group annotators. The x-axis shows the proportion of annotators within that group (in-group, red; out-group, blue) who flagged the tweet as hate speech. Each subplot contains 200 points across 100 tweets.}
    \label{fig:alignment}
\end{figure}

We hypothesized that one's desire to label online text as hate speech would change based on their underlying assumption about the group membership of the author o the text. Because human annotators were only given the text of the post and not other critical context such as the user's profile picture, follower network, or post engagement and reactions, we ask annotators to consider both cases of in- vs. out-group membership of the author (Q6 and Q7, respectively). We then calculate the observed probability for each text being flagged as hate speech by calculating the proportion of annotators answering `Yes' from the total annotations for that term ($n_{f-word}=6$, $n_{b-word}=9$, and $n_{n-word}=6$). 
Our results show that indeed, a user's attitude towards whether text should be flagged as hate speech changes based on the assumed group membership of the text's author. This is seen visually by the clustering of in-group and out-group labels along the x-axis in Fig. \ref{fig:alignment}. Aside from the b-word, we see that in-group points tend to have lower rates of hate speech labeling compared to out-group points. %

We also compare these probabilities with Perspective API's identity attack score, assuming that these scores are properly calibrated probabilities. Fig. \ref{fig:alignment} plots Perspective API's identity attack score against the proportion of annotators marking it as hate speech for each example ($n=100$ per slur). The observed probabilities for in-group (Q6) and out-group (Q7) are shown in red and blue, respectively. We additionally plot the line of best fit using the least squares polynomial fitting along with the Pearson correlation for each (denoted as R). With a range from -1 to 1, a higher R value indicates a higher correlation between Perspective API's score and annotator's proportion of hate speech labeling.
Regarding Perspective API's alignment with user's attitudes, our results show distinct patterns across slurs, suggesting disparate user experience. The f-word has no visual clusters, confirmed with the lowest Pearson correlation scores. The b-word, on the other hand, has two visual clusters. Additionally, the in-group correlation is the highest here, suggesting that Perspective API's identity attack score is best aligned with this community's attitudes towards the use of the b-word by other in-group community members. The n-word shows stark distinctions, in that Perspective API does not score a signal text below 0.40. In other words, text including the n-word is comparatively given a higher identity attack score. 
Our results suggest that Perspective API's model is disproportionally aligned to certain reclaimed slurs, namely the b-word, while more than likely disproportionally suppressing the speech of LGBTQIA+ and Black community members engaging in online reclaimed slur use. This uneven calibration mirrors cross-slur differences in human interpretation: the b-word appears more legible to both annotators and the model, while the n-word and f-word retain greater normative ambiguity, a pattern we examine explicitly in RQ4. 

To represent the visual spread seen in Fig. \ref{fig:alignment}, we calculate the average total variation (ATV), defined as

\vspace{-5pt}
\begin{align*}
    ATV_{\textrm{in/out-group}} &= \frac{1}{n} \sum_{i=1}^n |\Delta|, \text{where} \\
    \Delta &= P_{\textrm{human}}(t_i \textrm{ being hate speech} | \textrm{in/out-group}) - P_{\textrm{Perspective}}(t_i \textrm{ being an identity attack}),
\end{align*}
and $t_i$ is text $i$. Lower ATV is better, as it signifies more alignment between the distributions we're comparing.

\begin{table}[h]
\centering
\small
\begin{tabular}{lrr}
\toprule
 & $ATV_{\textrm{in-group}}$ & $ATV_{\textrm{out-group}}$ \\
 \midrule
f-word &  0.31 & 0.30 \\
b-word &  0.18 & 0.22 \\
n-word &  0.42 & 0.14 \\
\textbf{all words}  &  0.30 & 0.22 \\

\bottomrule
\end{tabular}
\caption{\label{tab:tav_results} Average total variation (ATV) results reported by group membership and reclaimed slur.}
\end{table}

Table \ref{tab:tav_results} displays these results. Overall, $ATV_{\textrm{in-group}}$ is  0.30 and $ATV_{\textrm{out-group}}$ 0.22, confirming that Perspective API's model is more calibrated to assume that a text's author is external to the group that has reclaimed that slur. The f-word has similar in-group and out-group ATV scores (0.31 and 0.30, respectively). The b-word has the lowest $ATV_\textrm{in-group}=0.18$, meaning Perspective API's model is best aligned to online users' attitudes around this slur. Interestingly, this mimics the comparative societal malleability of the b-word, which is now used broadly in mainstream venues. The n-word, on the other hand, displays the highest $ATV_\textrm{in-group}$ of 0.42 and lowest $ATV_\textrm{out-group}$ of 0.14. This signifies that Perspective API's model is calibrated to flagging online text as an identity attack, assuming that the text's author is not part of the Black community. Critically, these results validate online Black users' reported experiences of being suppressed in online spaces when using the n-word, and their subsequent need to self-censor to avoid that suppression \cite{Harris_2023, Chance_Chang_2025}. Crucially, these calibration gaps are not uniform across reclaimed slurs, underscoring that harms from automated moderation are unevenly distributed.

\begin{table}[t]
\footnotesize
\centering
\begin{tabularx}{\textwidth}{l c >{\raggedright\arraybackslash\hangindent=1.5em\hangafter=1}X}
\toprule

\textbf{f-word} & \textbf{$\Delta$} & \\
\midrule

\multirow{4}{*}{ingroup} & -0.72 & [MASK] [MASK] Gay gay f*ggy gay gay gay queery f*golishious gay gay gay gay \\

& -0.70 & Smoking 20 f*gs at once is insanity \\

& -0.62 & ``Durr n*gger. Durrr tr*nny. Durr f*g. Durr c*nt. Durr durr durr durrrrr look at me I'm a nazi haha doesnt this make me so quirky and a free thinker durrr''. I mean seriously you are fucked up individuals for talking this way \\

\midrule

\multirow{3}{*}{outgroup} & 0.932 & RT [MASK]: Muscle f*gs are prone to rough pounding \\

& 0.90 & [MASK] Shut up F*g bag \\

& 0.79 & [MASK] Ron Paul is based. Why are you all being f*gs about it? \\

\midrule
\textbf{b-word} & \textbf{$\Delta$}  & \\
\midrule

\multirow{9}{*}{ingroup} & -0.62 & Black baby ain't really gonna be great. And that's really for baby's teething, those chicken wings to raise them black Y'all really thought a b*tch couldn't rap, huh? Perhaps this your answer for that, a crack addict is the average black dad And n*ggas are still scared of, nah actually this is for me This one for TT at the dude serving the mac and the cheese  \\

& -0.55 & I'm a big black b*tch and I love it! \\

& 0.54 & [MASK] [MASK] [MASK] Better not see you b*tches crying when Kamala walks all over our constitutional rights because you felt Trump wasn't doing enough in the moment on 2A which could easily be negotiated. Short sighted. Dumb! \\

\midrule

\multirow{7}{*}{outgroup} & 0.73 & [MASK] [MASK] [MASK] B*tch, you haven't played sports a day in your life... \\

& 0.72 & actually texting her 20 times a day is still not enough she'll put up with it if you got money. The best way to drive that b*tch away is to let her know that you got no money and no future prospects \\

& 0.71 & I find it hilarious how she suddenly sees herself as some sort of crusader for people with endometriosis (If she even has it). Some people have this condition yet hold down a full time very busy very stressful job (like myself!). She is just an uptight b*tch looking for fame \\

\midrule
\textbf{n-word} & \textbf{$\Delta$} & \\
\midrule

\multirow{4}{*}{ingroup} & -0.74 & N*gga be drunk just like me and my n*ggas [loudly\_crying\_face] \\

& -0.73 & this n*gga DOUBLED DOWN???? \\

& -0.72 & If you don't wanna be my n*gga for life just don't talk to me i don't feel like getting to know someone and then it all to be over — wow, n*ggas have feelings too lol \\

\midrule

\multirow{3}{*}{outgroup} & 0.67 & Sisi ugly n*ggas hatunanga kioo Kwa mansion, we believe we look like God \\

& -0.50 & i hate being called a n*gga \\

& -0.38 & Are you able to picture being able to get non-n*gger-made KFC? No, because its made best by us n*ggas \\

\bottomrule
\end{tabularx}
\caption{\label{tab:topvar_examples}
Examples of the top variation ($\Delta$) observed between Perspective API identity-attack predictions and annotations, assuming in- vs. out-group author membership. Negative values indicate that Perspective API's score was greater than the human annotator's observed probability, whereas positive values indicate the annotator's observed probability was greater than Perspective API's. All slurs have been censored for the purpose of presenting these examples in the paper.}
\end{table}

To elucidate ATV difference, we extract the top 3 examples for Q and Q7 with the highest variation $\Delta$ (difference in human and API labeling) as seen in Table~\ref{tab:aggregated_tv}. Overall we find 3 major patterns. The first is repetition and term combination as seen in example 1, 3, 13, 15, and 18. Within these examples, the repetitive and overuse of one or more slurs tend to cause high likelihoods of identity attack presence for Perspective API, while for the human annotators, they are able to identify the context such as quoting hateful content or reclamatory use. The second pattern is alternative meaning as seen in example 2 in which the f-word is used as a neologism for cigarettes in British culture. As this is common knowledge in the queer community, while not necessarily common in the training data for Perspective API, this large $\Delta$ value is expected. Lastly, is lack of context as seen in example 7. In this example, the author of the tweet was paraphrasing song lyrics from the song "Self" by Noname. Perspective API lacks that context and therefore would label it based on the usage of the n-word. Together, these findings suggest that high $\Delta$ values are rarely arbitrary, they reflect systematic gaps between statistical pattern matching and the contextual reasoning that human annotation affords.

\subsection{RQ4: What patterns and differences emerge in how reclaimed slurs are interpreted across different marginalized communities?}

Taken together, our findings demonstrate the reclaimed slurs cannot be treated as a homogeneous category in moderation or evaluation. Differences across slurs reflect distinct social functions, historical context, and dependence on author identity, challenging approaches that assume reclamation can be captured through a single interpretive framework. We explore this further in Section \ref{sect:interview_insights}.

Across the corpus, reclaimed slurs frequently appeared within broader patterns of antagonistic language, particularly for the b-word, which often functioned as an amplifier rather than direct target, appearing in  expressions (e.g., ``son of a b*tch'') or intensifying insults aimed at other groups. In contrast, co-occurrences involving all three studied slurs were rare and typically accompanied by additional slurs outside the focal communities (e.g., t-word and c-word). These patterns highlight how reclaimed terms are embedded within broader systems of meaning rather than operating in isolation. These asymmetries are further exposed in comparison between human judgment and automated predictions. Models consistently favor interpretations that treat authors as out-group members, with the largest misalignment occurring for the n-word under assumed in-group authorship. This mirrors broader societal tendencies to deny or flatten reclamatory uses of language tied to historical and structural violence and reflects the limitation of modeling hate speech without the author identity. We underscore that we do not support predicting author identity to ``improve'' automated hate speech detection, as it supports surveillance infrastructure.

Overall, our results indicate that reclaimed language operates through slur-specific logic shaped by context, identity, and power. Moderation systems tailored to one reclaimed term may fail or even cause harm when applied to others. Effective moderation therefore requires approaches that recognize intra- and cross-community variation rather than collapsing reclaimed slurs into a single evaluative category.

 \subsection{Interview Insights} \label{sect:interview_insights}

To complement the quantitative findings, we draw on insights from semi-structured interviews as to understand how lived experiences, beliefs, and upbringing further influence a personas engagement and boundaries with reclaimed slurs. By analyzing responses thematically, we identify recurring trends and shared perspectives across participants.

Discussions of the b-word consistently surfaced cultural differences in meaning, acceptability, and harm. Interviewees ($n=2$) emphasized that dominant Anglo-American norms and broader alignment with whiteness in the U.S. shape how the term is interpreted and moderated $[P1]$. In several non-U.S. contexts, the word was described as being used more explicitly to uphold patriarchal norms and attack women $[P2]$, whereas in the U.S. it remains misogynistic but is also used more flippantly, particularly by women and queer communities $[P1]$. As one interviewee noted, 
\begin{quote}
    \textit{The n-word and f-word ... will both immediately get censored out of things where the b-word doesn't} -$P1$.
\end{quote}

Both interviewees described growing up primarily encountering the term in misogynistic contexts but later being exposed to reclaimed uses through college, social media, and broader community engagement. This shift complicated annotation practices, particularly in cases where context was sparse or ambiguous. One interviewee explained that, 
\begin{quote}
    \textit{Seeing it three or four times, I think these lines are talking about women of color specifically but it's written deadpan with no context ... this could be nothing or hate speech.} - $P1$

\end{quote}
\noindent Interviewees also noted generational differences in usage and acceptability, observing that older speakers tend to use the term primarily in derogatory ways, while younger speakers are more likely to employ it in a reclaimed manner. 
 
For the f-word, interviewees emphasized parallels with slurs targeting queer and trans communities in other languages. All three described prior exposure to related slurs within their native language, some of which were used for sexual empowerment or reclamation $[P3]$. Another interviewee noted that slurs across English and their native language reflect existing class and caste power structures, with the English f-slur perceived as more ``chic'' than native-language slurs $[P4]$. Interviewee $P4$ emphasized that within their cultural context, harm was not inherent to a specific word, noting that it could be ``any word really'', but instead depended on derogatory intent and usage.

Interviewees also mentioned that the f-word has been used as a catch-all slur due to limited sociocultural delineation between queer and trans identities. Two interviewees $[P5, P3]$ noted that this ambiguity shaped both harm and reclamation, particularly for trans individuals. One interviewee mentioned that, as a trans person, they did not reclaim the f-word, viewing it as more closely reclaimed by cis gay male communities and therefore as out-group language. As interviewee $P3$ explained,
\begin{quote}
    \textit{I think what ... makes it difficult is ... different people are just going to have a different emotional response.}
\end{quote}
They further reflected on the challenges this poses for moderation, stating, 
\begin{quote}
    \textit{A person's like saying that this is harmful and ... how much can you like even trust this person's assessment is correct ... if I were to remove this would like, you know, mitigating the harm for them be worth possibly also harming you know, the people who are using it for any reason. It's really complicated. I don't know what a better world would be like.}
\end{quote}
 This complexity informed the interviewee's skepticism and distrust toward automated moderation systems, particularly those operated by private companies. Interviewee $P3$ described an experience in which their Facebook Marketplace account was removed due to the presence of “trans” in their bio, reinforcing concerns that content moderation systems have historically aligned with platform interests rather than with broader goals of harm reduction or community care.

For the n-word, both interviewees described a shift in boundaries around the word that emerged through changes in community and increased engagement with its historical context. This process clarified their comfort levels and reinforced the view that acceptable use should be limited to in-group members. Interviewee $P6$ noted that in their upbringing in Nigeria, other slurs such as the f-slur were more commonly used, shaping early exposure and interpretation. Interviewee $P7$ raised concerns about community-guided moderation systems, noting how mechanisms like community notes can amplify false claims when enough users agree, and emphasized that current content moderation systems are not equipped to handle these nuances. They further compared this word to the b-word, explaining that while the n-word is often used in malicious intent, context remains critical, including in-group usage where other forms of subjugation, such as class, can still be present. 
Interviewee $P7$ also highlighted the distinction between the -a and-er forms of the n-word, noting that the latter more strongly signals malice, while the former is more commonly reclaimed, situating this linguistic shift within broader Black cultural practices.

\section{Implications and Conclusion}

Our results highlight that in-group annotation alone is insufficient to produce a reliable set of labels for reclaimed language. Content moderation systems and evaluation practices should treat disagreement not as error but as a meaningful reflection of plural interpretations within communities. As NLP increasingly turns toward pluralistic and community-aware approaches, our results underscore the need to rethink how representation is operationalized in annotation and moderation pipelines for language rooted in identity, history, and power.

\noindent\textbf{Broader Impact.} 
Our results suggests that in-group annotation, which is currently the standard solution, is not sufficient enough to produce an `accurate' set of labels. We have outlined the major limitations of this task, such as the inability to identify who (i.e. the community-based identities) is speaking/writing content, the inability to account for user's contextual lived experiences and beliefs around the slur, the inability to extract contextual cues and features from the text, as well as the implications of assuming author's group membership when moderating their content. Compounding this issue, we see very distinct trends across the three studied slurs. 
Current systems often treat features such as reclaimed language as term bias and swearing \cite{vidgen-etal-2021-learning, Salminen_Almerekhi_Milenkovi, francesconi-etal-2019-error}, grouping how systems handle reclaimed slurs into one method and ignoring the nuance of each slur. This inhibits our ability to decide whether their usage is reclamatory. This also inhibits these systems from learning what platforms users do and do not want to see, causing the system to guesses. While these systems are necessary, these systems do not have the capacity to fairly regulate reclaimed language. 

However, the more prominent question is, \textit{how should} reclaimed language be regulated? Across reclaimed slurs, there is lack of in-group community consensus on reclamation, inconsistent and historically and culturally biased social acceptability standards, and algorithmic alignment with dominant cultural power. Which communities' reclamation practices take precedence? How should context and intent be evaluated, and what mechanism ensure fairness without reinforcing digital colonialism?
Do we require in-group human-in-the-loop evaluation to assess each usage on platforms? 

\noindent\textbf{What does this mean for how communities interact with these systems?} Often, marginalized communities are responsible for protecting themselves, and in this case, the burden is no different. However, there are strategies that allow communities to navigate these systems more effectively. Existing work has shown the benefits of self-censorship techniques, such as asterisking or modifying words (e.g.,``shit'' $\rightarrow$ ``sh*t'') to avoid automated filtering \cite{Chance_Chang_2025}, as well as using algorithm-friendly alternatives or ``algo-speak'' to signal reclaimed language while bypassing sensitive moderation \cite{Calhoun_Fawcett_2023, doi:10.1177/20563051241254371}. Communities also leverage collective action to manage harmful content: as Interviewee $P6$ described, organizing local networks to mass report abusive accounts or posts can be an effective way to ensure platforms enforce rules against genuinely harmful behavior. These practices demonstrate that, while challenges remain, communities are finding ways to assert agency and preserve their linguistic and cultural spaces online.

\noindent\textbf{Where is content moderation heading?}
Content moderation systems are increasingly moving away from one-size-fits-all enforcement toward personalized and pluralistic approaches. A prominent direction is personalized content moderation (PCM), in which users can configure aspects of what content they see, often through adjustable thresholds for toxicity-based attributes or keyword-based filtering \cite{10.1145/3610080, gurkan2024personalized}. These systems promise greater user autonomy and, in some cases, improved user well-being \cite{moscato-etal-2025-personalization}.

However, this shift also redistributes  responsibility, moving moderation decisions from platforms to individuals. Prior work cautions that PCM may intensify filter bubbles, fragment shared norms, or permit content that conflicts with legal or community standards \cite{gurkan2024personalized, 10.1145/3610080, moscato-etal-2025-personalization}. Public discourse reflects these concerns, including opt-in settings that allow exposure to racism, xenophobia, or misogyny \cite{gault2021intel}. User studies further surface discomfort with ambiguous moderation attributes and the burden placed on individuals to manage harmful content. One participant highlighted that an ``n-word'' toggle is particularly problematic for African American users who use the term in reclaimed contexts, illustrating how de-contextualized personalization risks suppressing culturally specific expression \cite{10.1145/3610080}.

Parallel to PCMs is the discussion of pluralistic alignment: aligning AI systems to a plurality of human values and norms, rather than to a single consensus target, thereby enabling multiple valid interpretations or moderation outcomes \cite{pmlr-v235-sorensen24a}. Current approaches move away from universal enforcement toward community-centered moderation by surfacing a range of reasonable judgments, steering models to reflect specific norms, or aligning outputs with population-level distributions, with recent work demonstrating modular and low-resource methods for operationalizing these approaches \cite{pmlr-v235-sorensen24a,feng-etal-2024-modular,luo-etal-2025-towards}. As \citet{pmlr-v235-sorensen24a} note, “customization necessitates pluralism.” Future moderation systems may benefit from pluralistic alignment approaches that leverage smaller, community-specific models to guide platform decisions. Combined with deliberative alignment, which enforces policy-specific reasoning before decision are made \citep{guan2025deliberativealignmentreasoningenables}, these methods point towards more context-aware and community-centered moderation practices.

\section*{Positionality Statement} 
The authors of this paper belong to communities that have reclaimed the slurs that are centered in the paper.

\section*{Generative AI Usage Statement}
ChatGPT (GPT-5-mini) was used solely for grammar editing and to assist with debugging figure generation. No text in the manuscript was generated by AI. All writing, analysis, and conceptual contributions are the original work of the authors.

\nocite{*}
\bibliographystyle{ACM-Reference-Format}
\bibliography{references}

\appendix
\section{Additional Results and Analysis}

\begin{table}[H]
\footnotesize
\centering
\begin{tabular}{ l  r  r  r}
\toprule
     & \multicolumn{1}{c}{\textbf{\hspace{3mm}F-Word}} & \multicolumn{1}{c}{\textbf{\hspace{3mm}B-Word}} & \multicolumn{1}{c}{\textbf{\hspace{3mm}N-Word}}\\

     \cmidrule{2-4}

     & \multicolumn{1}{r}{N=600} & \multicolumn{1}{r}{N=900} & \multicolumn{1}{r}{N=600}\\
     \midrule

    \hspace{1mm}\textbf{Reclaimed Usage}& & \\
    \hspace{3mm}Insular & 242 & 194 & 350 \\
    \hspace{3mm}Pride & 52 & 101 & 89 \\
    \hspace{3mm}Neither & 300 & 530 & 161 \\
    \hspace{3mm}Missing & 6 & 75 & 0 \\[0.15cm]

    \hspace{1mm}\textbf{Salient Context}& & \\
    \hspace{3mm}Counter speech & 24 & 21 & 15 \\
    \hspace{3mm}Discussion of identity & 139 & 94 & 191 \\
    \hspace{3mm}Discussion of slur & 57 & 7 & 123 \\
    \hspace{3mm}Neologism & 70 & 198 & 88 \\
    \hspace{3mm}Quote & 34 & 38 & 71 \\
    \hspace{3mm}Recollection & 29 & 24 & 47 \\
    \hspace{3mm}Sarcasm & 34 & 41 & 33 \\
    \hspace{3mm}Sexualization & 131 & 239 & 12 \\
    \hspace{3mm}Missing & 82 & 238 & 20 \\[0.15cm]

    \hspace{1mm}\textbf{Secondary Salient Context}& & \\
    \hspace{3mm}Counter speech & 22 & 16 & 9 \\
    \hspace{3mm}Discussion of identity & 31 & 42 & 51 \\
    \hspace{3mm}Discussion of slur & 11 & 9 & 24 \\
    \hspace{3mm}Neologism & 12 & 52 & 9 \\
    \hspace{3mm}Quote & 6 & 6 & 16 \\
    \hspace{3mm}Recollection & 20 & 8 & 71 \\
    \hspace{3mm}Sarcasm & 10 & 9 & 33 \\
    \hspace{3mm}Sexualization & 12 & 26 & 8 \\
    \hspace{3mm}Missing & 476 & 732 & 379 \\[0.15cm]

    \hspace{1mm}\textbf{Derogatory}& & \\
    \hspace{3mm}No & 255 & 264 & 324 \\
    \hspace{3mm}Yes & 339 & 571 & 276 \\
    \hspace{3mm}Missing & 6 & 65 & 0 \\[0.15cm]

    \hspace{1mm}\textbf{Target}& & \\
    \hspace{3mm}A broader group of people & 63 & 143 & 68 \\
    \hspace{3mm}A known subset of people & 42 & 83 & 77 \\
    \hspace{3mm}Another individual & 103 & 287 & 165 \\
    \hspace{3mm}Individuals from a community associated with the slur & 254 & 126 & 239 \\
    \hspace{3mm}Individuals not from a community associated with the slur & 69 & 36 & 9 \\
    \hspace{3mm}Oneself & 41 & 98 & 42 \\
    \hspace{3mm}Missing & 28 & 127 & 0 \\[0.15cm]

    \hspace{1mm}\textbf{Ingroup Report}& & \\
    \hspace{3mm}No & 423 & 496 & 454 \\
    \hspace{3mm}Yes & 171 & 340 & 145 \\
    \hspace{3mm}Missing & 6 & 64 & 1 \\[0.15cm]

    \hspace{1mm}\textbf{Outgroup Report}& & \\
    \hspace{3mm}No & 258 & 366 & 273 \\
    \hspace{3mm}Yes & 335 & 471 & 325 \\
    \hspace{3mm}Missing & 7 & 63 & 2 \\[0.15cm]
    
    \bottomrule
\end{tabular}
\caption{\label{tab:annotation_label}Distribution of annotation labels across all reclaimed slurs for each annotation question. The F-Word had 6 annotator, B-Word had 9, and N-Word had 6. Each question sums to $\text{\# of annotator} \times 100$ per slur across the label options.}
\end{table}

\begin{center}
\begin{table}[t]
\footnotesize
\centering
\begin{tabular}{ l  r  r  r}
\toprule
     & \multicolumn{1}{c}{\textbf{\hspace{3mm}F-Word}} & \multicolumn{1}{c}{\textbf{\hspace{3mm}B-Word}} & \multicolumn{1}{c}{\textbf{\hspace{3mm}N-Word}}\\

    \cmidrule{2-4}
    & \multicolumn{1}{c}{\textit{\hspace{3mm}B (95\% CI)}} & \multicolumn{1}{c}{\textit{B (95\% CI)}} & \multicolumn{1}{c}{\textit{B (95\% CI)}}\\
    \toprule

    \hspace{1mm}\textbf{Reclaimed Usage} \textit{(ref = Neither)}& & \\
    \hspace{3mm}Insular & 0.19$^{***}$ (0.12, 0.27) & 0.12$^{***}$ (0.06, 0.19) & 0.04 (-0.05, 0.13) \\
    \hspace{3mm}Pride & 0.42$^{***}$ (0.28, 0.55) & 0.17$^{***}$ (0.08, 0.26) & 0.12$^{**}$ (0.01, 0.23) \\
    \hspace{3mm}Missing & --- & 0.08 (-0.09, 0.26) & --- \\[0.15cm]

    \hspace{1mm}\textbf{Salient Context} \textit{(ref = Discussion of identity)}& & \\
    \hspace{3mm}Counter speech & 0.03 (-0.15, 0.21) & -0.13 (-0.30, 0.04) & -0.11 (-0.32, 0.11) \\
    \hspace{3mm}Discussion of slur & -0.15$^{**}$ (-0.28, -0.02) & -0.10 (-0.38, 0.17) & 0.23$^{***}$ (0.136, 0.329) \\
    \hspace{3mm}Neologism & 0.15$^{**}$ (0.02, 0.28) & -0.14$^{***}$ (-0.23, -0.05) & 0.50$^{***}$ (0.40, 0.61) \\
    \hspace{3mm}Quote & -0.09 (-0.25, 0.07) & -0.14$^{**}$ (-0.28, -0.00) & 0.56$^{***}$ (0.44, 0.67) \\
    \hspace{3mm}Recollection & -0.09 (-0.27, 0.08) & -0.04 (-0.20, 0.12) & -0.09 (-0.22, 0.05) \\
    \hspace{3mm}Sarcasm & 0.11 (-0.04, 0.27) & -0.16$^{**}$ (-0.29, -0.03) & 0.12 (-0.04, 0.27) \\
    \hspace{3mm}Sexualization & 0.15$^{***}$ (0.05, 0.25) & -0.09$^{**}$ (-0.18, -0.00) & 0.18 (-0.06, 0.418) \\
    \hspace{3mm}Missing & 0.03 (-0.09, 0.15) & -0.12$^{**}$ (-0.21, -0.02) & 0.33$^{***}$ (0.14, 0.52) \\[0.15cm]

    \hspace{1mm}\textbf{Derogatory} \textit{(ref = No)}& & \\
    \hspace{3mm}Derogatory (Yes) & 0.15$^{***}$ (0.07, 0.23) & 0.06$^{**}$ (0.00, 0.12) & -0.04 (-0.11, 0.03) \\
    \hspace{3mm}Missing & --- & 0.93$^{***}$ (0.76, 1.10) & --- \\[0.15cm]

    \hspace{1mm}\textbf{Target} \textit{(ref = Another individual)}& & \\
    \hspace{3mm}A broader group of people & -0.08 (-0.22, 0.05) & -0.07$^{**}$ (-0.15, -0.00) & 0.09 (-0.03, 0.21) \\
    \hspace{3mm}A known subset of people & 0.02 (-0.13, 0.17) & -0.07 (-0.15, 0.02) & -0.02 (-0.13, 0.10) \\
    \hspace{3mm}Individuals from a community associated with the slur & 0.15$^{***}$ (0.05, 0.25) & 0.03 (-0.05, 0.10) & -0.05 (-0.13, 0.04) \\
    \hspace{3mm}Individuals not from a community associated with the slur & -0.07 (-0.20, 0.07) & -0.04 (-0.16, 0.08) & -0.11 (-0.39, 0.17) \\
    \hspace{3mm}Oneself & 0.05 (-0.12, 0.22) & -0.12$^{***}$ (-0.21, -0.04) & -0.01 (-0.15, 0.13) \\
    \hspace{3mm}Missing & -0.06 (-0.27, 0.15) & -0.15$^{**}$ (-0.27, -0.03) & --- \\[0.15cm]

     \hspace{1mm}\textbf{Annotator Variance} & 0.00 & 1.00 & 1.00 \\[0.15cm]

    \hspace{1mm}\textbf{Intercept} & 0.00 (-0.14, 0.14) & 0.22$^{***}$ (0.12, 0.32) & 0.10$^{*}$ (-0.01, 0.20) \\[0.15cm]

    \midrule
    Observations & \multicolumn{1}{r}{600 (group num. = 6)} & \multicolumn{1}{r}{900 (group num. = 9)} & \multicolumn{1}{r}{600 (group num. = 6)}\\
    \bottomrule
\end{tabular}
\caption{\label{tab:mixed-effect}Linear mixed-effects model with a binary dependent variable indicating whether an annotator’s reporting decision changes as a function of the author’s group membership. The outcome equals 1 when the annotator gave different judgments to the in-group and out-group reporting prompts, and 0 when the judgments were identical. The model includes random intercepts grouped by text instance and annotator, where the annotator-level variance reflects systematic differences in how individual annotators weigh group membership in their reporting decisions. Fixed effects include reclaimed-usage features, contextual categories, derogatory-use labels, and target-of-slur features. [key: $^{*}$p<0.1, $^{**}$p<0.05, $^{***}$p<0.01]}
\end{table}
\end{center}

\begin{figure}
    \centering
    \includegraphics[width=\linewidth]{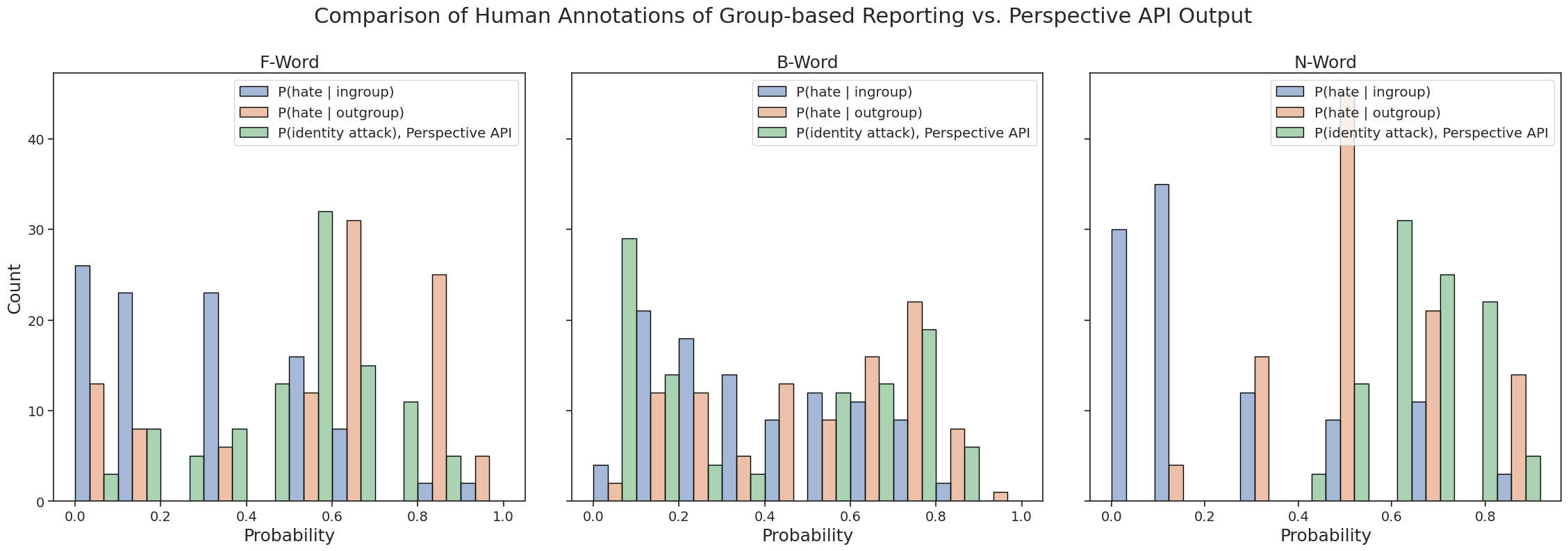}
    \caption{Distributional comparisons between the probability of a human annotator labeling the text as hate speech (conditioned on the assumption that it was written by someone in- vs. out- group) and Perspective API's predicted probability of it being an identity attack.}
    \label{fig:calibration}
\end{figure}

\begin{table}[t]
\centering
\footnotesize
\begin{subtable}{0.3\textwidth}
\centering
\caption{f-word}
\begin{tabular}{lcc}
\toprule
 & Out-group & In-group \\
\midrule
At-least-one & 0.21 & 0.28 \\
Majority     & 0.30 & 0.69 \\
\bottomrule
\end{tabular}
\end{subtable}
\hfill
\begin{subtable}{0.3\textwidth}
\centering
\caption{b-word}
\begin{tabular}{lcc}
\toprule
 & Out-group & In-group \\
\midrule
At-least-one & 0.36 & 0.36 \\
Majority     & 0.24 & 0.38 \\
\bottomrule
\end{tabular}
\end{subtable}
\hfill
\begin{subtable}{0.3\textwidth}
\centering
\caption{n-word}
\begin{tabular}{lcc}
\toprule
 & Out-group & In-group \\
\midrule
At-least-one & 0.07 & 0.27 \\
Majority     & 0.18 & 0.74\\
\bottomrule
\end{tabular}
\end{subtable}

\caption{Total variation distance between Perspective API outputs and human judgments, stratified by voting rule and group membership for text author.}
\label{tab:aggregated_tv}
\end{table}

\newpage

\section{Extended Literature Review} \label{appendix:literature_extended}

\textbf{Reclaimed Slurs.}
Slurs are pejorative terms targeting groups, often associated with race, ethnicity, nationality, religion, sexual orientation, gender, or disability in a U.S. context. Rooted in historical trauma, colonialism, and patriarchy, they have been used to oppress communities and elevate the caller's social status. Beyond colloquial use, slurs have been codified in law and politics, further perpetuating negative stereotypes. As these terms become socially taboo, they default to neutral descriptors.
However, target communities have increasingly reclaimed these words, using them positively to signal resilience, pride, and solidarity. Various models describe this process: \citet{BIANCHI201435}'s echo model distinguishes \textit{friendship} (non-political use) from \textit{appropriation} (deliberate socio-political subversion of the slur). \citet{Galinsky} outlines three levels of reclamation: individual in-group labeling, group adoption as a symbol, and out-group acceptance and revaluation. \citet{Reclamation} studies Black and queer communities, noting barriers to group-level reclamation due to self-stigmatization and proposing a payoff matrix that incorporates power dynamics. For this study, we leverage Jeshion's polysemy model \cite{jeshion}, which defines \textit{insular} reclamation (in-group camaraderie) and \textit{pride} reclamation (expressing dignity and honor in group membership).

\textbf{Automated Content Moderation Suppression.}
Content moderation systems disproportionately suppress marginalized users by mis-enforcing policies on identity-related speech. Case studies illustrate how moderation often constrains expression rather than protects communities. \citet{Harris_2023} show that Black creators on TikTok experience shadow banning and content removal when posting about Black identity under vague violations, with little transparency into algorithmic or moderation decisions. \citet{Shaid2023} characterize moderation as digital colonialism, showing that Western-centric norms on Facebook lead to inconsistent hate speech enforcement, deletion of posts or groups, censorship of historical content, and suppression of humor and political resistance, suppressing Bangladeshi Facebook users. These findings demonstrate systematic suppression of marginalized expression.

Research on moderation datasets and models further exposes structural limitations. Hate speech datasets exhibit racial bias, with language associated with Black users more likely to be labeled toxic or abusive \cite{davidson-etal-2019-racial}. Iterative dataset construction \cite{vidgen-etal-2021-learning} and implicit hate speech modeling across minority groups \cite{hartvigsen-etal-2022-toxigen} improve detection but treat harmful language as a classification problem, often overlooking context and lived experience. Reclaimed language poses particular challenges: models disproportionately flag non-derogatory LGBTQIA+ slur usage by gender-queer authors as harmful, even when identity context is provided \cite{dorn2024harmfulspeechdetectionlanguage}. Survey studies also reveal identity-based moderation disparities, with Black and trans users more likely to have content or accounts removed for discussing identity, justice, or critiques of dominant groups \cite{Haimson2021}. Collectively, these works show how both platform governance and automated systems reinforce dominant norms, suppressing marginalized voices.

Our work is inspired and influenced by the work done in \citet{dorn2024harmfulspeechdetectionlanguage}, this groundwork allowed us to expand upon the annotation methodology to other communities outside of the gender-queer community and look more closely into the annotation process. Specifically, our annotation and methodology approach expands on \citet{dorn2024harmfulspeechdetectionlanguage} to both explore salient context and usage of the slurs as well the implication of group membership of the author of the text.

\textbf{Reclaimed Language in NLP.}
Although hate speech detection has been widely studied, reclaimed language remains understudied. Most work addresses it indirectly through identity term bias, where slurs or identity markers increase the likelihood of content being labeled hateful or abusive regardless of context \cite{attanasio-etal-2022-entropy, 10.1145/3278721.3278729, sap-etal-2019-risk}. Keyword bias studies show texts containing slurs are more often flagged as offensive or abusive, with inconsistent definitions of offensive, abusive, and toxic across datasets compounding the issue \cite{YIN2022100210, cercas-curry-etal-2024-subjective}. Efforts to mitigate this include modeling subjectivity so that identity terms in ambiguous statements are not treated as strong harm signals \cite{zhao2021ssbertmitigatingidentityterms}, though these approaches focus on spurious correlations rather than understanding reclamation itself.

A smaller set of studies directly examines reclaimed language. \citet{kurrek-etal-2020-towards} propose a taxonomy of derogatory, reclaimed, and counter-speech uses of slurs on Reddit, highlighting annotation challenges and biases in Perspective API toward reclaimed or discussion-based uses. \citet{dorn2024harmfulspeechdetectionlanguage} introduce QueerReclaimLex, showing that hate speech classification models and LLMs disproportionately flag in-group LGBTQIA+ slur usage as harmful. Similarly, \citet{10.1145/3614419.3644025} create the Reclaimed Hate Speech Dataset (RHSD) and demonstrate that classifiers mislabel reclaimed LGBTQIA+ language as hateful, increasing false positives and risking suppression of self-expression. Other work fine-tunes models to predict reappropriation using homo-transphobic datasets \cite{draetta-etal-2024-reclaim}, but still frames reclamation as a binary classification task.
Overall, existing NLP work largely treats reclaimed language as a modeling problem rather than a socially situated practice shaped by disagreement, context, and lived experience. Our work builds on these efforts by centering reclaimed slur usage, examining intra-community disagreement, and challenging the assumption that consensus or identity-aware labels can fully capture reclamation.

\textbf{Annotator Selection.}
Prior work highlights the need for intentional, stratified selection of annotators for subjective content moderation tasks, foregrounding that positionality influences annotations \cite{santy-etal-2023-nlpositionality, fleisig-etal-2023-majority, 10.1145/3555088}. \citet{santy-etal-2023-nlpositionality} find high disagreement in social acceptability and hate speech judgments across demographic groups, highlighting the need to center harmed communities during data annotation and study design. \citet{larimore-etal-2021-reconsidering} consider in-group (non-white) vs. out-group (white) communities to analyze how lived experience bias annotations. However, even the comparison of in-group and out-group flattens and generalizes the identities and preferences of in-group members, assuming that all people of color contextualize and experience instances of hate similarly. We hypothesize that within communities, hate speech annotation agreement for text involving slurs is low.

\textbf{Annotator Disagreement.}
Annotator identity, including demographics, lived experiences, and beliefs, significantly shapes subjective labeling tasks such as toxicity and hate speech detection \cite{sap-etal-2022-annotators, goyal2022toxicitytoxicityexploringimpact, pei-jurgens-2023-annotator, biester-etal-2022-analyzing}. This variance manifests across multiple dimensions of rater identity: \citet{al-kuwatly-etal-2020-identifying} train models on the same data with distinct annotator groups and find significant performance differences along lines of first language, age, and education, while \citet{waseem-2016-racist} find that amateur annotators label more text as hate speech compared to experts, even when both groups receive training. Cross-cultural studies further highlight this instability such as \citet{lee-etal-2024-exploring-cross} who found substantial label disagreement across countries, driven by differing interpretations of sarcasm and personal biases on divisive topics. Compounding this, \citet{santy-etal-2023-nlpositionality} show that annotators with different backgrounds align with different models, and that datasets and models disproportionately reflect the perspectives of white, Western, college-educated annotators. Studies show that crowd annotators differ systematically from experts, and that disagreement is influenced by cultural, political, and experiential factors. Simple aggregation of labels or in-group versus out-group approaches \cite{fleisig-etal-2023-majority,10.1145/3555088} fail to capture the nuanced, heterogeneous perspectives within identity groups and are therefore insufficient to fully represent lived experiences and interpretive differences.

Recent work adopts pluralistic approaches to disagreement, modeling annotator differences via embeddings, or historical or demographic information \cite{deng-etal-2023-annotate,10.1609/aaai.v37i12.26698}, distribution-aware calibration \cite{baan-etal-2022-stop}, and modular pluralism frameworks for LLM alignment \cite{feng-etal-2024-modular,bansal2025comparingbadapplesgood}. These methods explicitly account for diverse perspectives across communities, producing models sensitive to heterogeneous judgments rather than enforcing a single ``gold standard.''
Such research underscores that annotator disagreement is not noise but reflects the socially situated nature of language interpretation. Intentional consideration of annotator identity combined with pluralistic methods enables NLP systems to better capture nuanced, subjective tasks, moving beyond simplistic consensus labels or in-group/out-group strategies.

\onecolumn

\section{Pre-Screening Survey}\label{appendix:prescreen_survey}

{\larger\textbf{Section 1. Annotator Screening Form for Content Moderation for Reclaimed Language}}\\

\noindent Hello!

\noindent We would like to invite you to participate in the research being conducted by Christina Chance, Arjun Subramonian, and Rebecca Pattichis as part of our graduate NLP research at UCLA. We are interested in how marginalized communities perceive and are affected by content moderation, in particular as it relates to English-language social media content using reclaimed language.

\noindent As part of this project, we ask members of Black, LGBTQIA+, and women communities to annotate social media comments that use reclaimed language for whether they constitute hate speech. We are also soliciting interviewees for qualitative interviews to contextualize decisions around whether content containing reclaimed language is hate speech. This is a screening form to solicit annotators and interviewees for this project. Our questions are shaped by an American diasporic context. 

\noindent The data from this form will be collected, and it will be stored until the end of the project (estimated December 2025), after which it will be deleted. Only the project leaders (listed above) will have access to the raw data and no raw data will be made public except in aggregate.

\noindent This form will take approximately 10 minutes to complete. 

If you have any question about the study, please feel free to reach out to Christina Chance (cchance@ucla.edu)!

\noindent If you have any question about the study, please feel free to reach out to \textit{first author's name and email}!
\\

\noindent \textbf{Informed Consent Agreement}

\noindent Please click and read the informed consent agreement below.

\noindent I have read and agree to the informed consent agreement.

\begin{itemize}
    \item[$\circ$] I agree
\end{itemize}

\vspace{5mm}

\noindent {\larger\textbf{Section 2. Compensation}}

\noindent This is a screening form. You will only be compensated if you complete the form AND are selected to participate in the annotation and/or Zoom interview. To be clear, you will only be compensated if you get selected to participate. 

\vspace{3mm}

\noindent If you qualify for the study, we will contact you with more information. The approximate rate is \$45 for 100 annotations and \$10 for an approximately 15-20 minutes Zoom interview. Payment will be given in the form of an Amazon gift card, and will be emailed to you after completion of the annotations and/or interview.

\begin{itemize}
    \item[$\circ$] Yes, I understand the terms of compensation
\end{itemize}

\vspace{5mm}

\noindent {\larger\textbf{Section 3. Criteria for Participation}}\\

\noindent \textbf{Are you 18 years of age or above? }
\begin{itemize}
    \item[$\circ$] Yes, I am
\end{itemize}

\vspace{1mm}

\noindent \textbf{Are you proficient in English?}
\begin{itemize}
    \item[$\circ$] Yes, I am
\end{itemize}

\vspace{1mm}

\noindent \textbf{Do you use any social media platform (e.g., Twitter/X, Facebook, Instagram, Youtube)?}
\begin{itemize}
    \item[$\circ$] Yes, I ado
\end{itemize}

\vspace{1mm}

\noindent \textbf{Are you a part of at least one of these communities: LGBTQIA+, Black, and/or woman?}

\noindent The content of this form will include reclaimed language from the LGBTQIA+, Black, and woman communities; hence, being part of at least one of these communities is a requirement for participation.
\begin{itemize}
    \item[$\circ$] Yes, I am
\end{itemize}

\vspace{1mm}

\noindent \textbf{Warning:  If selected for this study, you will potentially be exposed to hateful, discriminatory, and explicit textual content.}
\begin{itemize}
    \item[$\circ$] Yes, I understand
\end{itemize}

\vspace{5mm}

\noindent {\larger\textbf{Section 4. Background Information}}

\noindent These questions are motivated by our personal experiences with reclaimed language. We are deeply invested in understanding the nuanced perspectives of Black, LGBTQIA+, and women communities as researchers that come from these communities.

\noindent Towards this end, we will use your demographic information to study how perspectives on the use of reclaimed language in social media content diverge within and between communities. The demographic information will only be presented in aggregate, unless explicit consent is provided for it to be presented otherwise, and not be used for any other analyses. Only the project leaders will have access to your demographic information for privacy and security reasons. We recognize the trust required to provide us with your demographic data, and we are happy to address any questions or concerns that you have about this data collection process.

\vspace{3mm}

\noindent \textbf{Email} (\textit{Short Answer Response})\\

\noindent \textbf{Age} (\textit{Short Answer Response})\\

\noindent \textbf{Do you identify as Black?}
\begin{itemize}
    \item[$\circ$] Yes
    \item[$\circ$] No
\end{itemize}

\vspace{1mm}

\noindent \textbf{Race/Ethnicity/Nationality} (\textit{Short Answer Response})

\noindent We understand that race/ethnicity are a complex concept. Please list racial, ethnic, and/or national categories with which you belong to. Please separate with commas. We provide some examples below:

\noindent Mexican-American, Cypriot, White\\
\noindent African-American, Black\\
\noindent Caribbean-American, Jamaican, Black\\ 
\noindent South-Asian, Indian-American

\noindent \textbf{Are you a woman?} 
\begin{itemize}
    \item[$\circ$] Yes
    \item[$\circ$] No
\end{itemize}

\vspace{1mm}

\noindent \textbf{Which of the following pronouns would you like to use in this context?}
\begin{itemize}
    \item[$\Box$] She/her
    \item[$\Box$] He/him
    \item[$\Box$] They/them
    \item[$\Box$] Ze/zem
    \item[$\Box$] Xe/xem
    \item[$\Box$] Any/all
    \item[$\Box$] I am questioning my pronouns
    \item[$\Box$] Prefer not to answer
    \item[$\Box$] Other:
\end{itemize}

\vspace{1mm}

\noindent \textbf{Are you a member of the LGBTQIA+ community?}
\begin{itemize}
    \item[$\circ$] Yes
    \item[$\circ$] No
\end{itemize}

\vspace{1mm}

\noindent \textbf{Which of the following best describes your sexual orientation?}
\begin{itemize}
    \item[$\Box$] Gay
    \item[$\Box$] Lesbian
    \item[$\Box$] Bisexual
    \item[$\Box$] Pansexual
    \item[$\Box$] Asexual
    \item[$\Box$] Straight
    \item[$\Box$] Questioning
    \item[$\Box$] None of the above
    \item[$\Box$] Prefer not to answer
    \item[$\Box$] Other:
\end{itemize}

\vspace{1mm}

\noindent \textbf{Which of the following best describes your romantic orientation?}
\begin{itemize}
    \item[$\Box$] Gay
    \item[$\Box$] Lesbian
    \item[$\Box$] Bisexual
    \item[$\Box$] Pansexual
    \item[$\Box$] Asexual
    \item[$\Box$] Straight
    \item[$\Box$] Questioning
    \item[$\Box$] None of the above
    \item[$\Box$] Prefer not to answer
    \item[$\Box$] Other:
\end{itemize}

\vspace{1mm}

\noindent \textbf{Interview Participation}
\noindent The goal of the interview is to learn more about your personal views on the use of reclaimed language in your communities and to gather cultural and experiential instances to help understand how you use reclaimed language. We would contact you via the email address you have shared.
\begin{itemize}
    \item[$\circ$] Yes, I am willing to participate in an interview
    \item[$\circ$] No, I am not willing to participate in an interview
\end{itemize}

\vspace{1mm}

\noindent \textbf{Which word do you want to comment on? Please pick the word that you feel most comfortable or prefer commenting on/annotating if selected to participate.}

\noindent Even if you do not use this word but are a part of a community that does, we would still like your input. If you are a member of different groups, please choose a singular word to provide input on. If you are not a member of ANY of the communities that have ownership of the reclaimed word, please exit the form and do not continue. 
\begin{itemize}
    \item[$\circ$] N-word
    \item[$\circ$] F-Word (f-slur)
    \item[$\circ$] B-word (b*tch)
\end{itemize}

\vspace{5mm}

\noindent {\larger\textbf{Section 5. Reclaimed Usage of N-Word/F-Word/B-Word}}

\noindent This section will be based on the reclaimed use of the n-word. We will ask about your usage, in-group usage, out-group usage, and implications of the word. Please consider all of the following questions in the context of social media.

\vspace{2mm}

\noindent Proposed by Robin Jeshion in their 2020 paper, Pride and Prejudiced, are two predominant instances of reclaimed words:

\noindent\hangindent=1.5em\hangafter=1 \underline{Pride reclamation}: the use of slur as an expression of pride for being in-group, which is presented as an acceptable manner of referencing the group

\noindent\hangindent=1.5em\hangafter=1 \underline{Insular reclamation}:  the use of a slur as an expression of camaraderie amongst the members of a group, in which is presented as an unacceptable manner of referencing the group

\noindent \textbf{Do you consider this word to be reclaimed?}
\begin{itemize}
    \item[$\circ$] Yes
    \item[$\circ$] No
\end{itemize}

\vspace{1mm}

\noindent \textbf{Which communities do you think have reclaimed this word?} (\textit{Short Answer Response})

We recognized that people have different perspectives on which communities have reclaimed this word. When we refer to “marginalized group” in subsequent questions, we refer to the communities which you think have reclaimed it. \\

\noindent \textbf{Do you personally use this word in a reclaimed manner?}
\begin{itemize}
    \item[$\circ$] Yes
    \item[$\circ$] No
\end{itemize}

\vspace{1mm}

\noindent \textbf{Do you think others \textit{within} the marginalized group should use this word in a reclaimed manner?}
\begin{itemize}
    \item[$\circ$] Always
    \item[$\circ$] Sometimes
    \item[$\circ$] Never
\end{itemize}

\vspace{1mm}

\noindent \textbf{Do you think others \textit{outside} the marginalized group should be able to use this word?}
\begin{itemize}
    \item[$\circ$] Always
    \item[$\circ$] Sometimes
    \item[$\circ$] Never
\end{itemize}

\vspace{1mm}

\noindent \textbf{Do others \textit{you know outside} this marginalized group use this word?}
\begin{itemize}
    \item[$\circ$] Yes
    \item[$\circ$] No
\end{itemize}

\vspace{1mm}

\noindent \textbf{Should people within the marginalized group use this word to refer to people outside the marginalized group in a reclaimed manner?}
\begin{itemize}
    \item[$\circ$] Yes
    \item[$\circ$] No
\end{itemize}

\vspace{1mm}

\noindent \textbf{Should people within the marginalized group use this word in the company of people outside the marginalized group in a reclaimed manner?}
\begin{itemize}
    \item[$\circ$] Yes
    \item[$\circ$] No
\end{itemize}

\vspace{1mm}

\noindent \textbf{Please add any additional comments about word use such as the differences between different forms, the context/environment you find acceptable to use the word, etc. } (\textit{Long Answer Response})

\vspace{5mm}

\noindent {\larger\textbf{Section 6. Thanks for taking the time to fill out our screener!}}
\noindent We will email you if you have been selected to participate in this study. If you have any question about the study, please feel free to reach out to \textit{first author's name}!

\noindent \textbf{If you are part of more than one of the communities of interest, are there other words you are open to annotating?}
\begin{itemize}
    \item[$\Box$] N-word
    \item[$\Box$] F-word
    \item[$\Box$] B-word
\end{itemize}

\vspace{1mm}

\noindent \textbf{Please provide any feedback on your experience with the screening process.} (\textit{Long Answer Response})

\section{Annotation Questions}\label{appendix:annotation_questions}
During the creation of the annotation guidelines and questions, we performed five iterations of the task and clarified ambiguities of the instructions with other AI researchers within the FAccT and AIES communities. We provide further discussion on the purpose and framing around each annotation question below.

\begin{enumerate}
    \item [\textbf{Q1.}]\textbf{Suppose the author is in-group. What kind of reclamation is present in the use of the slur in this context?} For this question, we use the proposed framework of \citeauthor{jeshion} that defines two use cases of reclaimed language, and we include ``neither'' as a possible option to indicate that the slur is not being used in reclaimed manner. The words are defined below as they are defined in the annotation instructions provided.
    \begin{itemize}
        \item \textbf{Pride reclamation}: the use of slur as an expression of pride for being in-group, which is presented as an acceptable manner of referencing the group (e.g. queer and black).
        \item \textbf{Insular reclamation}: the use of a slur as an expression of camaraderie among the members of a group, in which it is presented as an unacceptable manner of referencing the group (e.g. n-word and b-word).
    \end{itemize}
    Often, context plays a large role in identifying whether a slur was used in a reclaimed manner, with essential context being whether the user identifies with a community that has reclaimed the word. Due to the nature of social media as well as limitations on the format of our data (i.e. lack of metadata or author social group identification), we cannot identify whether the speaker is in-group with certainty. While we could leverage the geo-location tagging associated with a user, bio information, or dialect identification to predict in-group membership, this could lead to erroneous conclusions about how a slur is being used. Moreover, many content moderation tasks requiring inferring whether text constitutes hate speech from the text alone. Furthermore, we believe that our in-group annotators often have the cultural and community insight to distinguish between genuine reclaimed uses of slurs and uses that intend to harm these communities. 
    \item [\textbf{Q2.}] \textbf{Do not make assumptions about whether the author is in-group or out-group. Is this a derogatory use of the word?} We chose to separate this from the question of whether the word is used in a reclamatory context, as we understand that any use of the slur by an individual, in-group or out-group, can still be derogatory (e.g., ``All you f*gs that got the tove lo tickets at 3 dollar bill in 0.2 seconds, may you perish''). We provided the definition of derogatory as defined in Webster dictionary: ``expressive of a low opinion: disparaging; detracting from the character or standing of something''. This question is a binary {\em yes} or {\em no} question. 

    \item [\textbf{Q3.}] \textbf{Whom is the slur directed at?} Slur use often varies across different target populations. Slurs may serve as a replacement for another word, be pluralized to represent a group of people within the reclaiming community, or even be used to describe a collective of both in- and out-group individuals. For this question, we provide the following options:  oneself, another individual, individuals from a community associated with the slur, individuals not from a community associated with the slur, a known subset of people, a broader group of people.

    \item [\textbf{Q4/5.}] \textbf{What is a salient context in which the word is being used?} We apply the slur usage taxonomy from \citep{kurrek-etal-2020-towards} and subcategories and definitions from \citep{dorn2024harmfulspeechdetectionlanguage} to categorize the usage type of each slur:

    \begin{itemize}
        \item \textbf{Counter Speech}: Response to an instance of derogation, in defense against a comment made by a single speaker or group.
    
        \item \textbf{Quote}: Reference to a slur embedded in a quote or paraphrase.
    
        \item \textbf{Discussion of Slur}: Discussion of a slur, its origin, or acceptable use cases.
    
        \item \textbf{Discussion of Identity}: Discussion of in-group identity dynamics and related concepts.
    
        \item \textbf{Sexualization}: Speaker uses slur to attribute sex or a sex role to others. 
    
        \item \textbf{Sarcasm}: A slur used ironically, contrary to its original meaning.
    
        \item \textbf{Recollection}: Recollection of a time a slur was used.
    
        \item \textbf{Neologism}: Slur contorted to a new linguistic format, such as using a noun as a verb or creating a new word entirely.
    
    \end{itemize}
    The secondary context question (Q5)  was not required.

    \item [\textbf{Q6/7.}] \textbf{Suppose the author is in-group/out-group. Would you want a content moderation model to report this as hate speech?} Independent of the determined usage context of the slur, annotators indicated whether they would consider the text to be hate speech. While people may understand the intent behind the usage of a slur, they may have different thresholds for the text to be automatically flagged as hate speech. We separate this question into an in-group and out-group version as group membership can highly influence whether the text consumer is comfortable seeing or interacting with the content. This question is a binary {\em yes} or {\em no} question. 
\end{enumerate}

\onecolumn

\section{Annotation Instructions}\label{appendix:annotation_instruct}

\noindent\textbf{Objective:} Classify the use of a slur in online content and assess whether the statement should be automatically detected as hate speech.\\

\noindent\textbf{Task Description:} Your task is to label online content according to predefined categories (described below). The overall goal of this annotation is to learn how \textbf{you}, based on \textbf{your lived experiences}, would label these examples. We are not treating your annotation as a representation of your identity but as a representation of how you understand the use of these slurs.  Note that it is okay for you to refer back to the annotation instructions while annotating, but we recommend that you DO NOT use the internet or other sources of information while completing the task.\\

\noindent Each row of the file corresponds to a different instance of online text containing a slur that you will be annotating for how it is used and whether you think it should be flagged by an automatic content moderation system. For each instance, there will be multiple questions, with dropdown options per question. Please follow the annotation instructions in this file for definitions of the categories in the form.\\

\noindent\textbf{Definitions:} We now provide definitions for the labels you will be categorizing online content into. The first definition comes from \href{https://www.merriam-webster.com/dictionary/derogatory}{Webster Dictionary}. The following two definitions are proposed by Robin Jeshion in their 2020 paper, Pride and Prejudiced
. The last set of definitions are provided by Rebecca Dorn in their 2024 paper, \href{https://dl.acm.org/doi/10.1145/3689904.3694704}{Harmful Speech Detection by Language Models Exhibits Gender-Queer Dialect Bias}.  If you have further questions, please feel free to reference the sources. \\

\begin{itemize}
    \item Derogatory: expressive of a low opinion: disparaging; detracting from the character or standing of something.
    \begin{itemize}
        \item ‘Derogatory’ and ‘offensive’ are not synonymous. The former centers the producer of the content, whereas the latter centers the impact of the content on the consumer. In-group members can say derogatory things to people in-group. 
    \end{itemize}
    \item Pride reclamation: the use of a slur as an expression of pride for being in-group, which is presented as an acceptable manner of referencing the group.
    \item Insular reclamation: the use of a slur as an expression of camaraderie amongst the members of a group, in which it is presented as an unacceptable manner of referencing the group.
    \item Counter Speech: Response to an instance of derogation, in defense against a comment made by a single speaker or group.
    \item Quote: Reference to a slur embedded in a quote or paraphrase.
    \item Discussion of Slur: Discussion of a slur, its origin, or acceptable use cases.
    \item Discussion of Identity: Discussion of in-group identity dynamics and related concepts.
    \item Sexualization:  Speaker uses slur to attribute sex or a sex role to others. 
    \item Sarcasm: A slur used ironically, contrary to its original meaning.
    \item Recollection: Recollection of a time a slur was used.
    \item Neologism: Slur contorted to a new linguistic format, such as using a noun as a verb or creating a new word entirely.
    \item Self Label: Speaker uses slur to reference themselves as a member of the ingroup. 
    \item Other Label: Slur ascribed to someone who is not the speaker. 
    \item Group Label: Slur used to describe a group of people.
    \item Reclamation: Slur use that places power with ingroup members.
\end{itemize}

\noindent\textbf{Annotation Questions}

\noindent You will be asked to label 100 tweets that contain a slur you chose in the survey screening. There are multiple questions you will be asked for each tweet with the following choices for answers:

\begin{enumerate}
    \item Suppose the author is in-group. What kind of reclamation is present in the use of the slur in this context?
    \begin{enumerate}
        \item Possible answers: pride, insular, neither
    \end{enumerate}
    \item Do not make assumptions about whether the author is in-group or out-group. Is this a derogatory use of the word?
    \begin{enumerate}
        \item Reminder: “Derogatory” is not synonymous with “offensive” or “harmful.” “Derogatory” can include mock impoliteness.
        \item Possible answers: yes, no
    \end{enumerate}
    
    \item Whom is the slur directed at? (multiple select)
    \begin{enumerate}
        \item Possible answer choices: oneself, another individual, individuals from a community associated with the slur, individuals not from a community associated with the slur, a known subset of people, a broader group of people
        \item An example of a known subset of people would be the writer of the content and their friends.
        \item In contrast, an example of a broader group of people could be women as a broad social category.
    \end{enumerate}

    \item What is a salient context in which the word is being used?
    \begin{enumerate}
        \item Possible answer choices: recollection, neologism, counter speech, quote, discussion of slur, discussion of identity, sexualization, sarcasm
    \end{enumerate}
    
    \item What is another salient context in which the word is being used? (if applicable)
    \begin{enumerate}
        \item Possible answer choices: recollection, neologism, counter speech, quote, discussion of slur, discussion of identity, sexualization, sarcasm
    \end{enumerate}
    
    \item Suppose the author is in-group. Would you want a content moderation model to report this as hate speech?
    \begin{enumerate}
        \item This question asks you, based on your personal experience with reading the content.
        \item Possible answer choices: yes, no
    \end{enumerate}
    \item Suppose the author is out-group. Would you want a content moderation model to report this as hate speech?
    \begin{enumerate}
        \item This question asks you, based on your personal experience with reading the content.
        \item Possible answer choices: yes, no\\
    \end{enumerate}

\end{enumerate}

\noindent\textbf{General Guidelines}
\begin{itemize}
    \item Read Carefully: Reach each comment thoroughly before assigning labels. Please perform the task to the best of your ability.
    \item Consistency: Be consistent with your labeling. Feel free to revisit and revise prior annotations.
    \item Contextual Specificity: Base your answers off of your lived experiences and reactions instead of answering as a representative of the various communities you belong to.
    \item Emotional Wellbeing: You will potentially be presented with toxic and hateful content; as such, please take wellness breaks as needed.
    \item External Information: Note that it is okay for you to refer back to the annotation instructions while annotating, but we recommend that you DO NOT use the internet or other sources of information while completing the task.
\end{itemize}

\section{Interview Questions} \label{appendix:interview_questions}

During our semi-structured ~15 -- 30 minute long interviews with a subset of the annotators, we asked the following questions:

\setlist{nolistsep}
\begin{enumerate}
    \item How do you identify in the context of the identities we are discussing today? Are there any intersections that you believe are impactful to this overall conversation?
    \item How did you experience those identities as you grew up and if they did, how did your experiences change over time? Can you tell us about your community and how they influenced your experiences over time?
    \item How has everything we have discussed (community and experiences over time) impacted your understanding and use of the reclaimed word (for which you annotated)?
    \item How do you feel about the reclaimed words being used in an unfamiliar space vs. a familiar space or with your community?
    \item How do you feel about reclaimed words being directed at you?
    \item Can you explain how you discovered and developed these rules around the use of the reclaimed word that you hold yourself to?
\end{enumerate}

\noindent Outside of these questions, we allowed interviewees to guide the conversation how they deemed most fit for their experiences and perspectives on the overall topic of reclaimed slurs and their experiences with content moderation systems.

\end{document}